\providecommand{\paperdocumentclass}{article}
\providecommand{\paperclassoptions}{11pt}
\documentclass[\paperclassoptions]{\paperdocumentclass}

\providecommand{\papernatbiboptions}{square,numbers,sort&compress}

\newif\ifpaperpreprintlayout
\ifdefined\papervenuebuild
  \paperpreprintlayoutfalse
\else
  \paperpreprintlayouttrue
\fi

\usepackage[T1]{fontenc}
\usepackage[utf8]{inputenc}
\ifpaperpreprintlayout
  
\usepackage[letterpaper,left=0.96in,right=0.96in,top=1.10in,bottom=1.05in,headheight=22pt,headsep=20pt,footskip=30pt]{geometry}
\usepackage{XCharter}
\usepackage[xcharter]{newtxmath}
\usepackage[scaled=0.95]{sourcesanspro}
\usepackage{microtype}
\usepackage{setspace}
\usepackage{caption}
\usepackage{fancyhdr}
\usepackage{needspace}
\usepackage{ragged2e}
\usepackage{titlesec}
\usepackage{xurl}

\fi
\usepackage{graphicx}
\usepackage{booktabs}
\usepackage{tabularx}
\usepackage{threeparttable}
\usepackage{array}
\usepackage{xcolor}
\usepackage{enumitem}
\usepackage{csquotes}
\usepackage{etoolbox}
\usepackage{subfiles}
\expandafter\usepackage\expandafter[\papernatbiboptions]{natbib}
\usepackage[acronym,nonumberlist,nopostdot]{glossaries-extra}
\usepackage[hidelinks]{hyperref}
\usepackage{orcidlink}
\usepackage[capitalise,noabbrev]{cleveref}

\graphicspath{{paper/figures/generated/}}
\setabbreviationstyle[acronym]{long-short}
\glsdisablehyper
\definecolor{draftred}{HTML}{A61B1B}

\usepackage{amsmath}

\usepackage{placeins}

\usepackage{pifont}

\newcommand{\paperbrand}{}
\newcommand{\paperversion}{}
\newcommand{\papertitle}{\textsc{TimeCap}: A Scalable Open-Source Pipeline for Captioning Wearable Health Time-Series Data}
\newcommand{\papersubtitle}{}
\newcommand{\paperpdftitle}{TimeCap: A Scalable Open-Source Pipeline for Captioning Wearable Health Time-Series Data}
\newcommand{\lastauthororcid}{0000-0002-8607-9148}
\newcommand{\lawrieorcid}{0000-0003-4192-9505}
\newcommand{\rosenblattlorcid}{0009-0007-6084-1441}
\newcommand{\deckorcid}{0009-0000-8604-487X}
\newcommand{\schuetzorcid}{0000-0001-5069-407X}
\newcommand{\kaarorcid}{0009-0008-9287-3138}
\newcommand{\millingorcid}{0000-0002-8842-2958}
\newcommand{\zhangorcid}{0000-0003-0850-4978}
\newcommand{\johnsonorcid}{0000-0002-1812-9553}
\newcommand{\jakoborcid}{0000-0003-4793-1366}
\newcommand{\schullerorcid}{0000-0002-6478-8699}
\newcommand{\fleischorcid}{0000-0002-4842-1117}
\newcommand{\kimorcid}{0000-0003-2971-1909}
\newcommand{\langerorcid}{0009-0000-6669-3928}
\newcommand{\xuorcid}{0000-0002-1017-3840}
\newcommand{\ashleyorcid}{0000-0001-9418-9577}
\newcommand{\paperauthors}{Max Rosenblattl\authororcid{\rosenblattlorcid}\textsuperscript{1}, Milana Gurbanova\textsuperscript{1}, Karl Aurel Deck\authororcid{\deckorcid}\textsuperscript{1}, Narayan Schuetz\authororcid{\schuetzorcid}\textsuperscript{1}, Thomas Kaar\authororcid{\kaarorcid}\textsuperscript{1,2,3}, Manuel Milling\authororcid{\millingorcid}\textsuperscript{4}, Nicolas Zumarraga\textsuperscript{3}, Ning Wang\textsuperscript{3}, Kelly W. Zhang\authororcid{\zhangorcid}\textsuperscript{5}, Anders Johnson\authororcid{\johnsonorcid}\textsuperscript{1}, Robert Jakob\authororcid{\jakoborcid}\textsuperscript{2,3}, Bj\"orn W.\ Schuller\authororcid{\schullerorcid}\textsuperscript{4,6}, Elgar Fleisch\authororcid{\fleischorcid}\textsuperscript{3,7}, Daniel Seung Kim\authororcid{\kimorcid}\textsuperscript{8}, Allan Lawrie\authororcid{\lawrieorcid}\textsuperscript{9}, Patrick Langer\authororcid{\langerorcid}\textsuperscript{1,2,3}, Maxwell A. Xu\authororcid{\xuorcid}\textsuperscript{10}, Euan Ashley\authororcid{\ashleyorcid}\textsuperscript{1}, and Paul Schmiedmayer\authororcid{\lastauthororcid}\textsuperscript{1}}
\newcommand{\paperpdfauthors}{Max Rosenblattl, Milana Gurbanova, Karl Aurel Deck, Narayan Schuetz, Thomas Kaar, Manuel Milling, Nicolas Zumarraga, Ning Wang, Kelly W. Zhang, Anders Johnson, Robert Jakob, Björn W.\ Schuller, Elgar Fleisch, Daniel Seung Kim, Allan Lawrie, Patrick Langer, Maxwell A. Xu, Euan Ashley, and Paul Schmiedmayer}
\newcommand{\paperaffiliations}{\textsuperscript{1}Stanford University, USA; \textsuperscript{2}Aionic Labs, Switzerland; \textsuperscript{3}Agentic Systems Lab, ETH Zurich, Switzerland; \textsuperscript{4}CHI, Technical University of Munich University Hospital, Germany; \textsuperscript{5}Imperial College London, UK; \textsuperscript{6}GLAM, Imperial College London, UK; \textsuperscript{7}Centre for Digital Health Interventions (CDHI), University of St.\ Gallen, Switzerland; \textsuperscript{8}University of Washington, USA; \textsuperscript{9}National Heart and Lung Institute, Imperial College London, UK; \textsuperscript{10}Google Research, USA}
\newcommand{\paperdate}{October 14, 2026}
\newcommand{\shorttitle}{TimeCap}

\newcommand{\correspondingemail}{mrosenbl@stanford.edu}
\newcommand{\paperrepositoryurl}{https://github.com/SchmiedmayerLab/TimeCAP}
\newcommand{\paperrepositorylabel}{github.com/SchmiedmayerLab/TimeCAP}
\newcommand{\papercodelicense}{MIT}
\newcommand{\paperdoiurl}{}
\newcommand{\paperdoilabel}{}
\newcommand{\paperkeywords}{}
\newcommand{\papersubject}{Preprint manuscript}

\newif\ifpaperdraft
\paperdraftfalse

\newcommand{\doi}[1]{\href{https://doi.org/#1}{doi:#1}}
\newcommand{\authororcid}[1]{\ifdefempty{#1}{}{\nobreak\hspace{0.22em}\orcidlink{#1}}}

\newcommand{\paperstartmaintext}{}
\newcommand{\paperkeywordsblock}{\ifdefempty{\paperkeywords}{}{\par\medskip
    \noindent\textbf{Keywords:} \paperkeywords
  }}
\newcommand{\paperbibliographystyle}{unsrtnat}
\ifSubfilesClassLoaded{\newif\ifpaperstandalonecitations
  \expandafter\pretocmd\csname NAT@sort@cites\endcsname{\global\paperstandalonecitationstrue}{}{}\pretocmd{\nocite}{\global\paperstandalonecitationstrue}{}{}\AtEndDocument{\ifpaperstandalonecitations
      \bibliographystyle{\paperbibliographystyle}\bibliography{references}\fi
  }}{}

\newcommand{\cmark}{\ding{51}}
\newcommand{\xmark}{\ding{55}}

\definecolor{statistical-orange}{RGB}{248,230,208}
\definecolor{structural-blue}{RGB}{218,228,243}
\definecolor{semantical-purple}{RGB}{216,210,231}
\definecolor{synthesizer-green}{RGB}{214,232,218}
\definecolor{statistical-orange-text}{RGB}{200,100,30}
\definecolor{structural-blue-text}{RGB}{40,90,180}
\definecolor{semantical-purple-text}{RGB}{110,70,160}
\definecolor{synthesizer-green-text}{RGB}{40,130,70}

\ifpaperdraft
  \input{paper/setup/draft-commands}
\fi

\newacronym{TSLM}{TSLM}{Time-Series Language Model}
\newglossaryentry{MHC}{name={\textsc{OpenMHC}},
  text={\textsc{OpenMHC}},
  sort={OpenMHC},
  description={open release of the My Heart Counts wearable dataset}}
\newacronym{WESAD}{WESAD}{Wearable Stress and Affect Detection}
\newacronym{VLM}{VLM}{Vision-Language Model}
\newacronym{WFM}{WFM}{Wearable Foundation Model}

\newacronym{LLM}{LLM}{Large Language Model}

\newacronym{HAR}{HAR}{Human Activity Recognition}
\newacronym{PPG}{PPG}{Photoplethysmography}
\newacronym{EDA}{EDA}{Electrodermal Activity}
\newacronym{IMU}{IMU}{Inertial Measurement Unit}
\newacronym{ECG}{ECG}{Electrocardiogram}
\newacronym{HRV}{HRV}{Heart Rate Variability}
\newacronym{HR}{HR}{Heart Rate}
\newacronym{AHMS}{AHMS}{Apple Heart and Movement Study}
\newacronym{AASM}{AASM}{American Academy of Sleep Medicine}
\newacronym{AFib}{AFib}{Atrial Fibrillation}
\newacronym{PD}{PD}{Parkinson's Disease}
\newacronym{CKM}{CKM}{Cardiovascular-Kidney-Metabolic}
\newacronym{CoT}{CoT}{Chain-of-Thought}
\newacronym{QA}{QA}{Question Answering}
\ifpaperpreprintlayout

\definecolor{paperink}{HTML}{2E2D29}
\definecolor{paperaccent}{HTML}{8C1515}
\definecolor{paperaccentdark}{HTML}{820000}
\definecolor{papermuted}{HTML}{53565A}
\definecolor{paperrule}{HTML}{C0C0BF}
\definecolor{paperwash}{HTML}{EAEAEA}

\renewcommand{\arraystretch}{1.14}
\setlist{leftmargin=*,topsep=0.5em,itemsep=0.22em,parsep=0pt,partopsep=0pt}

\titleformat{\section}
  {\bfseries\fontsize{15}{18}\selectfont\color{paperink}}
  {\thesection}
  {0.72em}
  {}
\titleformat{\subsection}
  {\bfseries\fontsize{12.2}{15}\selectfont\color{paperink}}
  {\thesubsection}
  {0.66em}
  {}
\titleformat{\subsubsection}
  {\bfseries\normalsize\color{paperink}}
  {\thesubsubsection}
  {0.62em}
  {}
\titleformat{\paragraph}[runin]
  {\bfseries\color{paperink}}
  {}
  {0pt}
  {}
  [.\quad]
\titlespacing*{\section}{0pt}{1.75\baselineskip plus 0.30\baselineskip minus 0.20\baselineskip}{0.52\baselineskip}
\titlespacing*{\subsection}{0pt}{1.20\baselineskip plus 0.20\baselineskip minus 0.15\baselineskip}{0.32\baselineskip}
\titlespacing*{\subsubsection}{0pt}{0.95\baselineskip plus 0.15\baselineskip minus 0.10\baselineskip}{0.24\baselineskip}
\titlespacing*{\paragraph}{0pt}{0.8\baselineskip}{0.5em}
\pretocmd{\section}{\Needspace{7\baselineskip}}{}{}
\makeatletter
\pretocmd{\subsection}{\if@nobreak\else\Needspace{7\baselineskip}\fi}{}{}
\makeatother

\newcommand{\preprintseparator}{\hspace{0.62em}\textcolor{paperrule}{\textperiodcentered}\hspace{0.62em}}
\newcommand{\preprintdateline}{\sffamily\fontsize{8.7}{10.5}\selectfont\color{papermuted}\ifdefempty{\paperbrand}{}{\paperbrand\preprintseparator}\ifdefempty{\paperversion}{}{\paperversion}\ifdefempty{\paperdate}{}{\ifdefempty{\paperversion}{}{\preprintseparator}\paperdate}}
\newif\ifpreprintmetadata

\renewcommand{\paperstartmaintext}{\clearpage}

\renewcommand{\maketitle}{\thispagestyle{preprintfirst}\vspace*{0.15in}\begingroup
  {\raggedright\bfseries\fontsize{25}{28.5}\selectfont\color{paperink}\papertitle\par}
  \ifdefempty{\papersubtitle}{}{\vspace{0.42em}
    {\raggedright\fontsize{13}{16}\selectfont\color{papermuted}\papersubtitle\par}}
  \vspace{0.95em}
  {\raggedright\bfseries\fontsize{11.1}{13.6}\selectfont\color{paperink}\paperauthors\par}
  \ifdefempty{\paperaffiliations}{}{\vspace{0.38em}
    {\raggedright\fontsize{9.2}{11.6}\selectfont\color{papermuted}\paperaffiliations\par}}
  \preprintmetadatafalse
  \vspace{0.64em}
  {\raggedright\sffamily\fontsize{8.45}{10.6}\selectfont\color{papermuted}\ifdefempty{\correspondingemail}{}{\textbf{Correspondence}\enspace\href{mailto:\correspondingemail}{\nolinkurl{\correspondingemail}}\preprintmetadatatrue
    }\ifdefempty{\paperrepositoryurl}{}{\ifpreprintmetadata\preprintseparator\fi
      \textbf{Code}\enspace\href{\paperrepositoryurl}{\nolinkurl{\paperrepositorylabel}}\preprintmetadatatrue
    }\ifdefempty{\papercodelicense}{}{\ifpreprintmetadata\preprintseparator\fi
      \textbf{Code license}\enspace\papercodelicense
      \preprintmetadatatrue
    }\ifdefempty{\paperdoiurl}{}{\ifpreprintmetadata\preprintseparator\fi
      \textbf{DOI}\enspace
      \href{\paperdoiurl}{\ifdefempty{\paperdoilabel}{DOI}{\nolinkurl{\paperdoilabel}}}\preprintmetadatatrue
    }\par
  }
  \endgroup
}

\renewenvironment{abstract}
  {\par\vspace{1.10em}
    \noindent\textcolor{paperrule}{\rule{\textwidth}{0.5pt}}
    \par\vspace{0.67em}
    \noindent{\bfseries\fontsize{11.2}{13.5}\selectfont\color{paperink}Abstract\par}
    \vspace{0.24em}
    \begingroup
    \small
    \setstretch{1.065}
    \setlength{\parindent}{0pt}
    \setlength{\parskip}{0.34em}}
  {\par
    \endgroup
    \vspace{0.58em}
    \noindent\textcolor{paperrule}{\rule{\textwidth}{0.5pt}}
    \par\vspace{1.08em}
  }

\renewcommand{\paperkeywordsblock}{\ifdefempty{\paperkeywords}{}{\par\vspace{0.58em}
    {\sffamily\footnotesize\color{papermuted}\textbf{\color{paperink}Keywords}\enspace\paperkeywords\par}}}

\fancypagestyle{preprint}{\fancyhf{}
  \fancyhead[L]{\sffamily\footnotesize\color{papermuted}\shorttitle}
  \fancyfoot[C]{\sffamily\footnotesize\color{papermuted}\thepage}
  \renewcommand{\headrulewidth}{0.4pt}
  
}
\fancypagestyle{preprintfirst}{\fancyhf{}
  \fancyhead[R]{\preprintdateline}
  \fancyfoot[C]{\sffamily\footnotesize\color{papermuted}\thepage}
  \renewcommand{\headrulewidth}{0.4pt}
  
}
\fancypagestyle{plain}{\fancyhf{}
  \fancyhead[L]{\sffamily\footnotesize\color{papermuted}\shorttitle}
  \fancyfoot[C]{\sffamily\footnotesize\color{papermuted}\thepage}
  \renewcommand{\headrulewidth}{0.4pt}
  
}
\renewcommand{\headrule}{\hbox to\headwidth{\color{paperrule}\leaders\hrule height \headrulewidth\hfill}}

\hypersetup{
  colorlinks=true,
  linkcolor=paperaccentdark,
  citecolor=paperink,
  urlcolor=paperaccentdark,
  pdfborder={0 0 0},
  pdftitle={\paperpdftitle},
  pdfauthor={\paperpdfauthors},
  pdfsubject={\papersubject},
  pdfkeywords={\paperkeywords}
}

\AtBeginDocument{\color{paperink}
  \pagestyle{preprint}
}

\AtBeginEnvironment{thebibliography}{\clearpage\thispagestyle{preprint}}

\fi

\title{\papertitle}
\author{\paperauthors}
\date{\paperdate}

\begin{document}

\maketitle

\begin{abstract}
Time-series data generated continuously across industrial, environmental, and especially health domains produce multimodal signals whose scale and heterogeneity make their processing and interpretation particularly challenging.
\glspl{TSLM} promise scalable interpretation of these signals, yet remain bottlenecked by a fundamental lack of text-annotated time series pairs that are signal-grounded, reproducible, and extensible across modalities and schemas.
Existing paired corpora are either closed or limited in scale and modality coverage.
Annotation with large language models could scale them up, but its outputs cannot be regenerated exactly, which prevents others from re-deriving, extending, or auditing the resulting corpus.
To address this, we release \textsc{TimeCap}, an open deterministic captioning pipeline with three reusable extractor tiers and a cross-channel synthesis layer that operates directly on the signal with no language model in the loop, extensible to new schemas by writing channel definitions alone.
This positions \textsc{TimeCap} as a reproducible mechanism for future dataset generation rather than a one-off release.
Applying \textsc{TimeCap} to \gls{MHC}, the MyHeart Counts wearable dataset, yields \textsc{MHCap}, the largest open caption layer for multimodal wearable time-series data.
\textsc{MHCap} spans 19 Apple HealthKit channels and 7{,}005{,}763 captions over 908{,}944 person-days.
To showcase the impact of the corpus, we train \textsc{OpenTSLM}, a time-series language model, on \textsc{MHCap} and benchmark on health-outcome prediction.
Caption-augmented training (\textsc{OpenTSLM-Cap}) outperforms a label-only baseline on 29 of 32 per-user binary evaluations across 8 tasks.
\paperkeywordsblock
\end{abstract}

\glsresetall

\paperstartmaintext

\section{Introduction}

From factory machines and power grids to the human body, multivariate time series arise wherever systems are monitored over time.
The volume and heterogeneity of these data make their processing and interpretation particularly challenging.
Consumer health is one important example, where large volumes of wearable data can shape how people perceive and manage their health.
Cardiovascular disease, diabetes, and dementia are major contributors to global disease burden, particularly in older adults~\citep{GBD2019causes}.
Consumer wearables already capture signals that show early promise for tracking a wide range of conditions~\citep{schuetz2026openmhcacceleratingsciencewearable, erturk2025behavioral, adams2024watchpd}.
Given the volume of these data, realizing this promise at population scale requires robust, automated interpretation of these signals.
\Glspl{TSLM} promise to enable exactly this: models that can describe temporal patterns, support downstream prediction, and uncover structure in complex multivariate data~\citep{openTSLM}.
Yet, realizing this potential requires models that can reason over these signals, and such models face three compounding gaps that jointly obstruct progress.

First, the \textbf{data gap}.
The largest paired caption and time-series corpus for multimodal consumer wearables to date comprises 59.7M~hours from 103{,}731 participants but remains closed~\citep[][Table~1]{sensorLM2025}.
The largest \emph{open} alternative is roughly four orders of magnitude smaller in recorded hours and limited to smartphone- and smartwatch-derived modalities, totaling 5{,}648 question--answer pairs over 60 subjects of ExtraSensory recordings~\citep[][§3.1, §4]{sensorQAbenchmark}.

Second, the \textbf{methods gap}.
Existing pipelines either bring language models into the annotation loop, as in TimeOmni-1's hybrid pipeline, in which only 2{,}339 of 23{,}605 samples are human-curated~\citep[][§3.2]{timeomni1}, or replace real paired data altogether with synthetic generation, as in ChatTS~\citep[][§3]{chatts}.
Recent benchmarks further show that state-of-the-art LLMs still fall short on health time-series reasoning~\citep{hearts2026, xu2026ecg}.

Third, the \textbf{extensibility gap}.
Existing pipelines, such as the closed pipelines discussed above and \citet{sensorLM2025}'s hierarchical captioning, are tied to specific sensor sets and modality schemas; adapting them to a new wearable dataset typically requires redesigning the annotation pipeline rather than reconfiguring it, which limits the field's ability to grow open paired data alongside new devices.

We address all three gaps by introducing \textsc{TimeCap}, an open and deterministic captioning pipeline built from four composable extractor tiers (statistical, structural, semantic, and cross-channel) that operate directly on the signal and extend to any new wearable dataset.
We also introduce \textsc{MHCap}, an open paired-caption layer of the \gls{MHC} data.
\Gls{MHC} comprises wearable data~\citep{schuetz2026openmhcacceleratingsciencewearable} from the MyHeart Counts Cardiovascular Health Study, a smartphone-based study of physical activity, sleep, and cardiovascular health~\citep{mcconnell2017myheartcounts, Hershman2019}.
\textsc{MHCap} is created by applying \textsc{TimeCap} to its 19 HealthKit channels spanning iPhone activity, Apple Watch physiology, sleep stages, and ten workout categories.
With \textsc{TimeCap}, anyone with access to the signals can regenerate the captions.
As a result, the corpus (i) can grow without silent distribution shift, (ii) stays independent of the models evaluated on it, whereas LLM-written captions partly turn evaluation into distillation of the captioning model, and (iii) can be audited, since every caption traces back to a rule and threshold that can be changed and regenerated consistently.

The pipeline code is released under the MIT license, positioning \textsc{TimeCap} as a reusable resource for future dataset generation rather than a one-off release.
To accelerate research and spare others from regenerating the corpus, we will also release the generated captions on Harvard Dataverse.

\begin{itemize}[leftmargin=*]
    \item \textbf{Open, deterministic captioning pipeline.}
    We release a fully reproducible MIT-licensed captioning pipeline that produces captions deterministically from the signal, thereby enabling anyone with access to the signals to regenerate the corpus exactly.
    The pipeline introduces four modular extractor tiers (statistical, structural, semantic, and cross-channel) and is extendable to new wearable datasets via a channel-schema abstraction rather than requiring redesign.
    The first three tiers parallel \citet[][§3.2]{sensorLM2025}'s closed pipeline; the cross-channel tier composes per-channel evidence such as steps, distance, heart rate, and energy into multi-channel activity and sleep narratives that no open pipeline we are aware of provides.

    \item \textbf{Open caption layer for multimodal wearable time-series data.}
    The corpus contains 7{,}005{,}763 captions over 908{,}944 retained person-days from 2{,}081 sharable \gls{MHC} participants.
    Building on \gls{MHC}~\citep{schuetz2026openmhcacceleratingsciencewearable} (11,894 participants, ${>}60$ million hours, 19 Apple HealthKit channels), \textsc{MHCap} complements its signal data with the paired-text layer needed to train \glspl{TSLM} on large-scale consumer-wearable data, spanning all 19 HealthKit channels.
    This breadth of channel and modality coverage aligns with the needs of multi-modal health AI systems and is unavailable in any open paired corpus to date.
    To show that the underlying pipeline generalizes beyond consumer wearables, we additionally apply \textsc{TimeCap} to the \glspl{WESAD} lab-physiology dataset~\citep{wesad2018}, producing \textsc{WESADCap} with only dataset-specific channel definitions changed.

    \item \textbf{Evaluation using SoTA \glspl{TSLM}.} We train \textsc{OpenTSLM}~\citep{openTSLM} on \textsc{TimeCap} generated captions and benchmark on
    ~health-outcome prediction over the \gls{MHC} evaluation suite \citep{schuetz2026openmhcacceleratingsciencewearable}.
    The caption-augmented \gls{TSLM} outperforms an architecturally identical label-only baseline on the majority of health-outcome tasks, providing direct evidence that the captions confer training signal beyond label supervision alone.

\end{itemize}

\section{Related Work}

\subsection{Captioning Pipelines for Sensor Time-Series}
Approaches to aligning sensor time-series with natural language span a spectrum from deterministic, signal-grounded generation to large-scale language model pipelines.
At the deterministic end, template-based natural language generation converts structured sensor features into reproducible text through predefined rules~\citep{reiter2000building, gatt2018survey}.
As dataset scale demands have grown, the field has moved progressively toward the generative end, with recent work ingesting sensor readings as tabular inputs or pre-derived summaries into language models~\citep{health-llm, khasentino2025personalhealth, liu2024fewshot}, generating paired data through question-answering frameworks tailored to activity recognition~\citep{sensorQAbenchmark}.
Each step toward the generative end trades signal grounding and reproducibility for coverage, with surveys documenting systematic hallucination as a key challenge for annotation quality~\citep{tan2024llmannotation}.

SensorLM~\citep{sensorLM2025} is the closest prior work and the direct inspiration for our pipeline, demonstrating that template-based captioning at population scale is feasible through a three-tier hierarchy in which \textit{statistical} captions embed per-channel summary statistics, \textit{structural} captions detect trends and anomalies via sliding-window analysis, and \textit{semantic} captions associate activity and sleep events with timestamps~\citep[][§3.2]{sensorLM2025}.
The extensibility gap shared by SensorLM and comparable prior work arises as each pipeline is designed around a fixed sensor schema, so that applying it to a new wearable dataset requires redesigning rather than reconfiguring the annotation pipeline.
Concurrent work, SleepLM~\citep{sleeplm2026}, validates that hierarchical captioning generalizes to clinical domains, yet its pipeline remains confined to polysomnography and its training corpus is closed due to data access restrictions imposed by the underlying cohorts.

\textsc{TimeCap} addresses both gaps by adopting the template tradition's grounding and reproducibility, extending SensorLM's three-tier hierarchy with a cross-channel tier that composes per-channel evidence into unified multi-channel narratives, and introducing a channel-schema abstraction so that the same pipeline applies to any wearable dataset by writing channel definitions alone. \textsc{MHCap} is its first application at pretraining scale.

\ifpaperpreprintlayout\enlargethispage{\baselineskip}\fi
\subsection{Open paired Sensor-Language wearable time-series data}

While large-scale wearable datasets spanning more than 100{,}000 participants already exist~\citep{doherty2017ukbiobank, weber2024nako, patten2026allofus}, the landscape of paired textual annotations for wearable sensor data remains scarce.
The handful of datasets that move toward language grounding each expose a different limitation.
CAPTURE-24 provides structured activity labels alongside 3{,}883 hours of wrist accelerometry from 151 participants~\citep{chan2024capture24}, but its annotations are taxonomic codes rather than natural-language captions suitable for language model training.
Sensor-QA~\citep{sensorQAbenchmark} takes a further step by providing human-authored question-and-answer pairs over 60 subjects of ExtraSensory recordings, making it the closest open benchmark for language grounding, but its narrow scope in both subject count and modality coverage limits its utility as a large-scale pretraining corpus.
TSR-Suite~\citep{timeomni1} pairs an LLM analyzer, human reviewers, and an LLM rewriter to curate only 2{,}339 of its 23{,}605 reasoning samples~\citep[][§3.2]{timeomni1}, while ChatTS~\citep{chatts} replaces real paired data altogether with attribute-based synthetic time series and LLM-generated question-and-answer pairs~\citep[][§3]{chatts}.
Both achieve scale at the cost of grounding and reproducibility, and neither can be re-run bit-for-bit, as their LLM-driven steps are non-deterministic across runs.
The closest prior work, SensorLM~\citep{sensorLM2025}, demonstrates that automatic captioning at population scale is feasible, yet,  its corpus and pipeline remain proprietary and closed to the research community.
SleepLM~\citep{sleeplm2026}, demonstrates that template-based captioning is achievable in a clinical setting, its training corpus, however, cannot be released due to data access restrictions imposed by the underlying NSRR cohorts, and its pipeline is designed exclusively around sleep-specific signals and does not generalize to new sensor schemas.

\begin{table}[t]
  \centering
  \small
  \setlength{\tabcolsep}{4pt}
  \caption{Paired sensor and language corpora at a glance.
  \textsc{MHCap} combines an open caption corpus with deterministic generation, schema extensibility, and an open-source pipeline.
  \textbf{Open}: corpus and pipeline code released.
  \textbf{Det.}: caption generation is deterministic.
  \textbf{Ext.}: captioning pipeline extends to new sensor schemas via
  configuration rather than redesign.
  Numbers are reported as in the cited papers.}
  \label{tab:paired-corpora}
  \resizebox{\textwidth}{!}{\begin{tabular}{llllccc}
    \toprule
    \textbf{Corpus} & \textbf{Scale} & \textbf{Modalities} &
    \textbf{Annotation} & \textbf{Open} & \textbf{Det.} & \textbf{Ext.} \\
    \midrule
    SensorLM\textsuperscript{a}  & 59.7M h, 103{,}731 ppl         & PPG, ACC, EDA, TEMP, ALT          & 3-tier                 & \xmark & \cmark & \xmark \\
    Sensor-QA\textsuperscript{b} & 5{,}648 QA, 60 ppl             & IMU, audio, phone state           & human-written          & \cmark & \xmark & \xmark \\
    TSR-Suite\textsuperscript{c} & 23{,}605 (2{,}339 h.)          & multi-domain TS                   & LLM + human-in-loop    & \xmark & \xmark & \xmark \\
    ChatTS\textsuperscript{d}    & 134K samples (synthetic only)                  & univariate/multivariate TS        & LLM + synthetic        & \cmark & \xmark & \xmark \\
    SleepLM\textsuperscript{e}   & {>}100K h, {>}10K ppl          & PSG, (EEG/EOG, ECG, resp., EMG)               & multi-level captioning & \xmark & \cmark & \xmark \\
    \midrule
    \textbf{\textsc{MHCap}}      & 7.0M cap, 909k days, 2{,}081 ppl & 19 HealthKit channels           & 3-tier + cross-channel & \cmark\textsuperscript{f} & \cmark & \cmark \\
    \bottomrule
  \end{tabular}}
  \\[0.4em]
  \footnotesize
  \textsuperscript{a}\citet{sensorLM2025}.
  \textsuperscript{b}\citet{sensorQAbenchmark}.
  \textsuperscript{c}\citet{timeomni1}.
  \textsuperscript{d}\citet{chatts}.
  \textsuperscript{e}\citet{sleeplm2026}.
  \textsuperscript{f}Pipeline code is public; corpus access is described in Appendix~\ref{app:data_access}.
\end{table}

Table~\ref{tab:paired-corpora} summarizes these resources and makes clear that no existing corpus simultaneously achieves openness, deterministic generation, and extensibility to new sensor schemas.
Against this backdrop, \textsc{MHCap} addresses this gap by providing 7{,}005{,}763 deterministically generated captions over 908{,}944 person-days, spanning all 19 Apple HealthKit channels including activity, physiology, sleep stages, and ten workout categories.
\textsc{MHCap} provides an open, reproducible, pretraining-scale paired caption layer for multimodal consumer wearable time series.

\section{\textsc{TimeCap} and Corpora}
\label{sec:TimeCap_and_Corpora}

\begin{figure}[h]
  \centering
  \includegraphics[width=\linewidth,alt={TimeCAP pipeline from wearable sensor traces through channel standardization and four captioning tiers to aligned text annotations and metabolic prediction tasks.}]{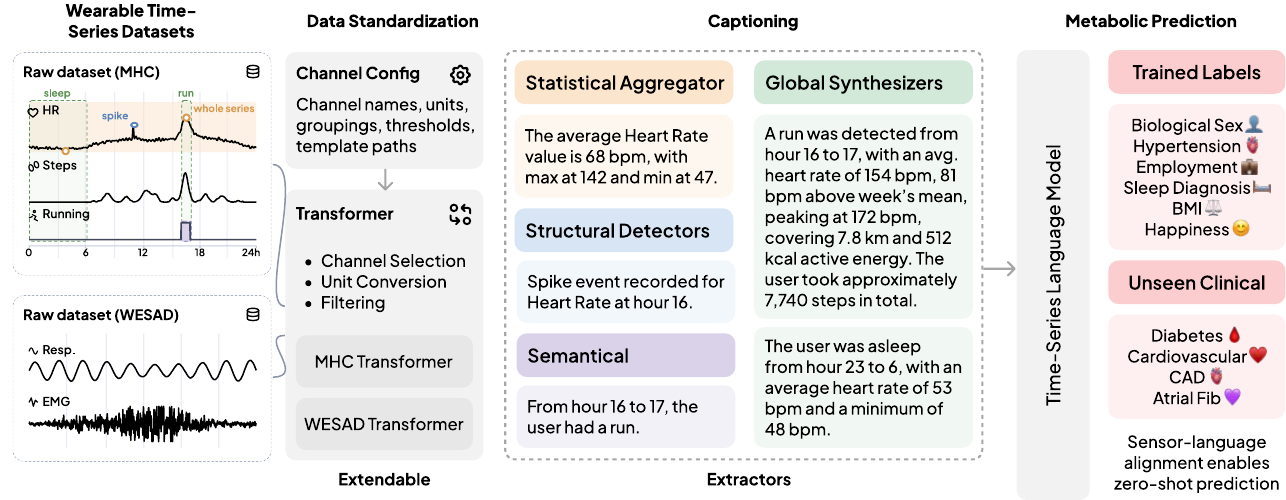}
  \caption{\textsc{TimeCap} pipeline overview.
  Raw multivariate sensor streams are mapped into a unified \texttt{Recording} representation, then captioned by four composable extractor tiers (statistical, structural, semantic, cross-channel) that emit a shared annotation schema.
  Cross-channel synthesis composes per-channel evidence into multi-channel narratives.}
  \label{fig:timecap-overview}
\end{figure}

We introduce \textsc{TimeCap}, a deterministic, dataset-agnostic captioning pipeline for converting multivariate time-series recordings into grounded natural-language captions.
Existing captioning approaches are typically dataset-specific and built on closed methodologies, limiting both reproducibility and extensibility.
\textsc{TimeCap} addresses both by replacing language-model generation with deterministic extractors operating directly on the signal, and by exposing a channel-schema abstraction so that the pipeline applies to any multivariate time-series data through channel definitions alone, rather than pipeline redesign.
Figure~\ref{fig:timecap-overview} provides a visual overview.

\paragraph{Schema}
Raw heterogeneous sensor streams are mapped into a unified representation by dataset-specific transformations, including choices such as channel selection, temporal resolution (e.g., minute- or hour-level aggregation), missing-value handling, alignment to a common time axis, and signal-specific preprocessing.
This step standardizes inputs into a shared \texttt{Recording} format, which stores the data as a channel-by-time matrix together with channel metadata such as names, display names, units, and data-availability summaries.
Missing values are encoded directly in the signal matrix, allowing downstream components to operate on a consistent interface independent of the original dataset schema.
This unified representation also supports textual annotations through a shared \texttt{Annotation} schema.
Each annotation records natural-language text, channel references, time intervals, and label as well as optional numerical values.

\subsection{Captioning}
We employ a modular per-channel set of extractors that transform standardized multivariate time series into a structured set of grounded signal-level captions.
Following prior work on hierarchical sensor-language alignment~\citep{sensorLM2025}, we decompose descriptions into four complementary levels, which capture increasing levels of abstraction from raw signals:

 \paragraph{\textcolor{statistical-orange-text}{Statistical Extractor}} Computes one summary annotation per continuous channel and recording, reporting four descriptive statistics: mean, maximum, minimum, and standard deviation.
 Each annotation is rendered through one of several deterministic natural-language templates.
 Channel-specific aggregation handles sensor-specific edge cases, for example excluding zero-valued heart-rate entries to avoid bias from inactive periods.
 The tier therefore yields at most one statistical caption per continuous channel, i.e., at most seven per \gls{MHC} week and exactly 16 per \gls{WESAD} session (Table~\ref{tab:corpus-scale}).

\paragraph{\textcolor{structural-blue-text}{Structural Extractor}} Localizes events within each channel, including spikes via \texttt{scipy.signal.\allowbreak find\_peaks}~\citep{SciPy2020} with channel-specific constraints on height, prominence, threshold, and separation.
Detects trends corresponding to sustained upward or downward changes over contiguous intervals by scanning the signal with multiple window sizes and strides, fitting a linear model within each candidate window, and retaining only those segments that satisfy predefined criteria on minimum span, coverage, effect magnitude, and goodness-of-fit.
Candidate segments are labeled as increasing or decreasing according to the fitted slope; nearby segments with the same direction can be merged into longer intervals, and overlapping or contained segments are resolved to form coherent trend annotations.

\paragraph{\textcolor{semantical-purple-text}{Semantic Extractor}} Extracts interval-based annotations directly from designated semantic channels such as sleep or workout activity channels that encode high-level event labels, typically in binary form.
The extractor identifies contiguous active intervals, filters out segments shorter than a configured minimum duration, and converts each retained interval into an explicit time-bounded annotation describing the activity state (e.g., "running", "sleeping").

\paragraph{\textcolor{synthesizer-green-text}{Cross-Channel Extractor}} Combines information across multiple signals to produce coherent natural-language annotations.
Rather than describing each channel in isolation, this stage reasons over temporal co-occurrence patterns and cross-channel relationships, allowing it to link high-level activity or sleep events with concurrent quantitative evidence from other modalities.
Where a semantic activity label is available, the cross-channel tier augments it with synchronized quantitative evidence from other modalities and aggregates statistics over the activity's full duration.
A labeled walking interval, for example, is enriched with concurrent heart rate, step count, distance covered, and active energy (see Figure~\ref{fig:SENSORCap-walking1}).
Similarly, sleep-related synthesizers can distinguish periods spent in bed from periods spent asleep by combining the corresponding HealthKit sleep channels, including intervals where the user is in bed but not asleep.
Conceptually, these annotations are defined by relationships between signals rather than by properties of any single channel alone.
Cross-channel captions provide a holistic multi-modal description of behavior that no per-channel tier produces in isolation, addressing the cross-modal narrative gap left open by prior closed pipelines.

All extractor hyperparameters are fixed before annotation generation, and the entire pipeline is deterministic with no language model in the generation loop.
Examples of generated annotations are provided in Appendix \ref{app:generated_captions}.

\subsection{\textsc{MHCap}}
Beyond introducing the \textsc{TimeCap} pipeline, we introduce \textsc{MHCap}, an open captioned dataset for multimodal consumer wearable time-series data.
Applied to the 19 Apple HealthKit channels of \gls{MHC}~\citep{schuetz2026openmhcacceleratingsciencewearable}, \textsc{MHCap} converts the underlying wearable traces into joinable daily and weekly caption layers for sensor-language pretraining.
These channels combine continuous HealthKit-derived measurements, including iPhone and Apple Watch mobility signals, heart rate, and active energy, with binary activity-state channels that mark semantically meaningful intervals such as sleep, walking, running, cycling, yoga, and strength-oriented activity.
The corpus contains 7{,}005{,}763 captions over 908{,}944 retained person-days from 2{,}081 sharable \gls{MHC} participants.
Each caption record retains the original \texttt{user\_id} and date, allowing captions to be matched back to the corresponding \textsc{OpenMHC} signal records.
Table~\ref{tab:corpus-scale} summarizes the resulting corpus scale and the distribution of captions across extractor tiers.

\begin{table}[h]
  \centering
  \small
  \caption{\textsc{TimeCap} corpus scale on \gls{MHC}-weekly and \gls{WESAD}.
  For \gls{MHC}-weekly, 7{,}005{,}763 captions are generated with a mean of 48.5 captions per user-week for a total of 2{,}081 participants.
  For \gls{WESAD}, 1{,}461 captions are generated with a mean of 97.4 captions per session for a total of 15 participants.}
  \label{tab:corpus-scale}
  \resizebox{\textwidth}{!}{\begin{tabular}{ccc|cccc}
    \toprule
    \multicolumn{3}{c|}{\textbf{Corpus}}
    & \multicolumn{4}{c}{\textbf{Captions by tier (count, share)}} \\
    \cmidrule(lr){1-3} \cmidrule(lr){4-7}
    \textbf{Dataset} & \textbf{Units\textsuperscript{a}} & \textbf{Sequences\textsuperscript{b}}
    & \textcolor{statistical-orange-text}{\textbf{Statistical}}
    & \textcolor{structural-blue-text}{\textbf{Structural}}
    & \textcolor{semantical-purple-text}{\textbf{Semantic}}
    & \textcolor{synthesizer-green-text}{\textbf{Cross-channel}} \\
    \midrule
    \gls{MHC}-weekly & 908{,}944 & 144{,}412
    & {1{,}002{,}819 \;(14.3\%)}
    & {3{,}191{,}600 \;(45.6\%)}
    & {867{,}948 \;(12.4\%)}
    & {1{,}943{,}396 \;(27.7\%)} \\
    \gls{WESAD} & 15 & 15
    & {240 \;(16.4\%)}
    & {1{,}071 \;(73.3\%)}
    & {75 \;(5.1\%)}
    & {75 \;(5.1\%)} \\
    \bottomrule
  \end{tabular}}
  \\[0.3em]
  \parbox{\textwidth}{\footnotesize
  \textsuperscript{a} Person-days (\gls{MHC}-weekly) or sessions (\gls{WESAD}).
  \textsuperscript{b} Weekly recordings ($168{\times}19$ tensors) or sessions (\gls{WESAD}).}
\end{table}

Dataset-specific channel mappings, detector thresholds, cross-channel synthesizer definitions, and template banks are provided in Appendix~\ref{app:snsrcap_config}.
A selection of generated captions for \gls{MHC} can be seen in Appendix \ref{app:snsrcap_captions}

\subsection{\textsc{WESADCap}}
To show that \textsc{TimeCap} generalizes beyond consumer wearable recordings, we additionally apply it to the controlled \gls{WESAD} stress and affect dataset~\citep{wesad2018}, producing \textsc{WESADCap}.
\textsc{WESADCap} captions 20 chest- and wrist-worn physiological channels, including accelerometry, electrodermal activity, temperature, respiration, blood-volume pulse, electromyography, and heart rate, together with binary labels for the baseline, stress, and amusement conditions and the meditation phases.
The same statistical, structural, semantic, and cross-channel tiers are instantiated at the session level, yielding captions that pair affective-state intervals with concurrent physiological and movement evidence.
\textsc{WESADCap} therefore complements \textsc{MHCap} with densely sampled multimodal physiology recordings grounded in affective-state context.
Table~\ref{tab:corpus-scale} summarizes the resulting corpus scale and caption-tier distribution.

Dataset-specific channel mappings, detector thresholds, cross-channel synthesizer definitions, and template banks are provided in Appendix~\ref{app:wesadcap_config}.
The captions generated by \textsc{TimeCap} for \gls{WESAD} can be seen in Appendix \ref{app:WESADcap_captions}

\section{Experiments and Evaluation}
\label{sec:results}

We evaluate \glspl{TSLM} on a downstream metabolic prediction task, predicting 10 participant-level health outcomes (8 binary, 2 ordinal) from a week of hourly-aggregated multi-channel sensor data.
The targets are selected from the \textsc{OpenMHC} label set~\citep{schuetz2026openmhcacceleratingsciencewearable} with the highest positive-class prevalence in the test split.
They are self-reported through in-app surveys~\citep{schuetz2026openmhcacceleratingsciencewearable}, so models trained on them are not intended for diagnosis or for decisions about individuals.
Each input is a $19 \times 168$ tensor over the 19 HealthKit channels at hourly resolution (24 hours $\times$ 7 days).
Six in-distribution targets (BiologicalSex, Hypertension, Employment, Sleep Diagnosis, BMI, Happiness) are seen during fine-tuning by both models, testing whether captions improve labels the model was supervised on.
Four held-out clinical conditions (Diabetes, Cardiovascular Disease, Coronary Artery Disease, Atrial Fibrillation) are excluded from fine-tuning and evaluated zero-shot, testing whether sensor-language alignment transfers to labels neither model has seen.

\paragraph{Baselines}
We compare two models that share the \textsc{OpenTSLM-SP} (soft prompting) architecture~\citep{openTSLM}, instantiated on the 1B and 3B variants of Llama-3.2~\citep{llama32}, but differ in training mix.
\textsc{OpenTSLM} is trained only on the metabolic outcome labels, while \textsc{OpenTSLM-Cap} additionally trains on structural, semantic and cross-channel captions from \textsc{MHCap} to encode text-time-series alignment.
We deliberately omit statistical captions as the time-series prompt already includes per-channel summary statistics.
Any performance gap between the two therefore reflects the caption supervision, not architectural differences.
We evaluate both models at the 1B and 3B parameter scales.
Training details and prompt templates are reported in Appendix~\ref{app:training_details}.

\paragraph{Metrics}
We report binary outcomes with macro-F1 and accuracy at the per-week level (Table~\ref{tab:opentslm_comparison}) and area under the ROC curve (AUROC) and area under the precision recall curve (AUPRC) at the per-user level (Table~\ref{tab:opentslm_comparison_user}).
Ordinal outcomes are reported with Spearman rank correlation ($\rho$) and quadratic weighted kappa (QWK) at both granularities.
Per-user AUROC, AUPRC, and Spearman~$\rho$ serve as the primary metrics, since the targets are static participant-level attributes.
Per-week metrics offer a finer-grained view of where caption training shifts predictions.
All metrics are computed on the held-out test split, with participant-level partitioning to prevent subject-level leakage.

\subsection{Results}
\label{sec:metabolic_task}

\paragraph{User-level performance}
On per-user AUROC and AUPRC (Table~\ref{tab:opentslm_comparison_user}), our caption-and-label finetuned \textsc{OpenTSLM-Cap} outperforms \textsc{OpenTSLM} on 29 of 32 binary metric-scale tasks across both 1B and 3B models.
At the user level, captions add a signal that persists through fine-tuning and helps on every binary task, whether or not the model saw it during training.

\begin{table}[h!t]
    \scriptsize
    \centering
    \caption{\textbf{User-level results.} \textsc{OpenTSLM-Cap} vs.\ \textsc{OpenTSLM} at the 1B and 3B parameter scales on the test split.
    Binary tasks report AUROC and AUPRC computed over per-user mean predictions.
    Ordinal tasks report Spearman~$\rho$ over the per-user mean predicted class, and QWK against the rounded per-user mean.
    The Majority column reports a class-prior baseline (always predict the majority class) as a no-skill floor and is excluded from best-of-row bolding.
    Best per task within each parameter scale in bold. $^{\dagger}$Held-out clinical task, not seen during fine-tuning.}
    \vspace{-0.5em}

    \setlength{\tabcolsep}{3pt}
    \renewcommand{\arraystretch}{1.05}

    \resizebox{\textwidth}{!}{\begin{tabular}{lcc@{\hskip 1.5em}cccc@{\hskip 1.5em}cccc}
    \toprule
        & \multicolumn{2}{c}{Majority} & \multicolumn{4}{c}{\textbf{1B}} & \multicolumn{4}{c}{\textbf{3B}} \\
        \cmidrule(lr){2-3} \cmidrule(lr){4-7} \cmidrule(lr){8-11}
    \textbf{Task}
        & \multicolumn{2}{c}{}
        & \multicolumn{2}{c}{\textsc{OpenTSLM-Cap}}
        & \multicolumn{2}{c}{\textsc{OpenTSLM}}
        & \multicolumn{2}{c}{\textsc{OpenTSLM-Cap}}
        & \multicolumn{2}{c}{\textsc{OpenTSLM}} \\
    \cmidrule(lr){4-5} \cmidrule(lr){6-7} \cmidrule(lr){8-9} \cmidrule(lr){10-11}
    \multicolumn{1}{l}{\textit{Binary (AUROC / AUPRC)}}
        & AUROC & AUPRC & AUROC & AUPRC & AUROC & AUPRC & AUROC & AUPRC & AUROC & AUPRC \\
    \midrule
    Biological Sex          & 50.00 & 25.38 & \textbf{79.24} & \textbf{51.72} & 78.90 & 49.99 & \textbf{77.86} & \textbf{51.24} & 75.47 & 49.80 \\
    Cardiovascular Disease$^{\dagger}$  & 50.00 & 15.30 & \textbf{65.15} & \textbf{29.99} & 62.09 & 26.79 & \textbf{67.19} & \textbf{28.94} & 65.28 & 26.25 \\
    Coronary Artery Disease$^{\dagger}$ & 50.00 & 6.90  & \textbf{65.88} & 15.76 & 63.38 & \textbf{16.62} & \textbf{62.46} & \textbf{12.57} & 59.94 & 9.69 \\
    Atrial Fibrillation$^{\dagger}$     & 50.00 & 3.23  & \textbf{68.36} & \textbf{7.24} & 59.82 & 4.91 & \textbf{64.08} & \textbf{7.33} & 61.72 & 5.40 \\
    Diabetes$^{\dagger}$    & 50.00 & 8.24  & \textbf{64.56} & \textbf{12.60} & 57.19 & 10.56 & 59.44 & \textbf{12.06} & \textbf{59.56} & 11.39 \\
    Employment              & 50.00 & 23.92 & \textbf{66.30} & \textbf{41.52} & 64.74 & 38.49 & \textbf{67.99} & \textbf{42.36} & 64.28 & 36.84 \\
    Hypertension            & 50.00 & 32.26 & \textbf{69.89} & \textbf{48.76} & 67.96 & 48.01 & \textbf{70.48} & \textbf{54.89} & 69.09 & 54.86 \\
    Sleep Diagnosis         & 50.00 & 19.06 & \textbf{59.52} & 25.35 & 58.86 & \textbf{26.13} & \textbf{60.43} & \textbf{26.53} & 57.95 & 24.99 \\
    \midrule
    \multicolumn{1}{l}{\textit{Ordinal (Spearman $\rho$ / QWK)}}
        & $\rho$ & QWK & $\rho$ & QWK & $\rho$ & QWK & $\rho$ & QWK & $\rho$ & QWK \\
    \midrule
    BMI (5 levels)          & 0.00 & 0.00 & 61.12 & 60.56 & \textbf{62.30} & \textbf{61.50} & \textbf{63.94} & \textbf{62.30} & 59.19 & 58.59 \\
    Happiness (4 levels)    & 0.00 & 0.00 & 8.71  & 4.00  & \textbf{13.30} & \textbf{6.57}  & \textbf{12.65} & \textbf{8.34} & 5.42 & 0.80 \\
    \bottomrule
    \end{tabular}}
    \label{tab:opentslm_comparison_user}
    \vspace{-1em}
\end{table}

\paragraph{Generalization}
Per-week macro-F1 (Table~\ref{tab:opentslm_comparison}) reveals where this advantage is concentrated.
Held-out clinical conditions show clear caption gains at the 3B scale: Cardiovascular Disease (+3.4), Diabetes (+6.1), Coronary Artery Disease (+2.6), and Atrial Fibrillation (+2.6).
At the 3B scale, in-distribution ordinal tasks also show clear caption gains: \textsc{OpenTSLM-Cap} improves QWK by +4.8 on BMI and +5.0 on Happiness over \textsc{OpenTSLM}.

\begin{table}[h!t]
    \scriptsize
    \centering
    \caption{\textbf{Per-week results.} Same model lineup at the (user, week) granularity, computed across all weekly predictions in the test split.
    Binary tasks (top) report macro-F1 (mean of class-0 and class-1 F1, label-symmetric; reported as F1) and accuracy; ordinal tasks (bottom) report Spearman~$\rho$ and QWK.
    The \textbf{Majority} column reports a class-prior baseline (always predict the majority class) as a no-skill floor and is excluded from best-of-row bolding.
    Best per task within each parameter scale in bold. $^{\dagger}$Held-out clinical task, not seen during fine-tuning.
    User-level metrics are given in Table~\ref{tab:opentslm_comparison_user}.}
    \vspace{-0.5em}

    \setlength{\tabcolsep}{3pt}
    \renewcommand{\arraystretch}{1.05}

    \resizebox{\textwidth}{!}{\begin{tabular}{lcc@{\hskip 1.5em}cccc@{\hskip 1.5em}cccc}
    \toprule
        & \multicolumn{2}{c}{\textbf{Majority}} & \multicolumn{4}{c}{\textbf{1B}} & \multicolumn{4}{c}{\textbf{3B}} \\
        \cmidrule(lr){2-3} \cmidrule(lr){4-7} \cmidrule(lr){8-11}
    \textbf{Task}
        & \multicolumn{2}{c}{}
        & \multicolumn{2}{c}{\textsc{OpenTSLM-Cap}}
        & \multicolumn{2}{c}{\textsc{OpenTSLM}}
        & \multicolumn{2}{c}{\textsc{OpenTSLM-Cap}}
        & \multicolumn{2}{c}{\textsc{OpenTSLM}} \\
    \cmidrule(lr){4-5} \cmidrule(lr){6-7} \cmidrule(lr){8-9} \cmidrule(lr){10-11}
    \multicolumn{1}{l}{\textit{Binary (F1 / Acc)}}
        & F1 & Acc & F1 & Acc & F1 & Acc & F1 & Acc & F1 & Acc \\
    \midrule
    Biological Sex                       & 43.47 & 76.91 & 64.59 & \textbf{78.71} & \textbf{65.20} & 78.12 & \textbf{64.87} & 78.29 & 61.24 & \textbf{78.40} \\
    Cardiovascular Disease$^{\dagger}$   & 43.49 & 76.94 & \textbf{61.65} & 72.62 & 61.13 & \textbf{76.37} & \textbf{62.00} & \textbf{70.22} & 58.62 & 67.13 \\
    Coronary Artery Disease$^{\dagger}$  & 47.10 & 89.02 & 58.83 & 76.02 & \textbf{61.11} & \textbf{81.91} & \textbf{55.60} & \textbf{70.78} & 52.96 & 66.82 \\
    Atrial Fibrillation$^{\dagger}$      & 49.24 & 97.02 & 43.68 & 67.76 & \textbf{44.99} & \textbf{71.41} & \textbf{44.39} & \textbf{67.97} & 41.79 & 62.20 \\
    Diabetes$^{\dagger}$                 & 46.99 & 88.63 & \textbf{53.14} & \textbf{67.60} & 51.35 & 65.91 & \textbf{59.12} & \textbf{74.07} & 53.02 & 67.89 \\
    Employment                           & 41.71 & 71.57 & \textbf{66.50} & \textbf{75.79} & 64.17 & 74.55 & 64.97 & 74.40 & \textbf{66.84} & \textbf{75.44} \\
    Hypertension                         & 36.86 & 58.37 & \textbf{64.56} & \textbf{66.54} & 64.01 & 65.67 & \textbf{65.52} & \textbf{68.38} & 64.34 & 66.46 \\
    Sleep Diagnosis                      & 44.21 & 79.25 & 55.68 & \textbf{75.11} & \textbf{56.81} & 74.57 & \textbf{57.01} & \textbf{75.24} & 56.98 & 74.68 \\
    \midrule
    \multicolumn{1}{l}{\textit{Ordinal (Spearman $\rho$ / QWK)}}
        & $\rho$ & QWK & $\rho$ & QWK & $\rho$ & QWK & $\rho$ & QWK & $\rho$ & QWK \\
    \midrule
    BMI (5 levels)                       & 0.00 & 0.00 & 59.21 & 60.18 & \textbf{59.47} & \textbf{60.70} & \textbf{61.59} & \textbf{62.40} & 56.31 & 57.62 \\
    Happiness (4 levels)                 & 0.00 & 0.00 & \textbf{7.79} & \textbf{7.54} & 6.80 & 5.04 & \textbf{9.02} & \textbf{8.02} & 4.70 & 3.00 \\
    \bottomrule
    \end{tabular}}
    \label{tab:opentslm_comparison}
    \vspace{-1em}
\end{table}

\FloatBarrier
\paragraph{Caption-family ablation}
We also train the 1B \textsc{OpenTSLM-Cap} on each caption family alone, with the same splits, filtering and schedule (Table~\ref{tab:caption_family_ablation}).
Any single family raises mean per-user AUROC over \textsc{OpenTSLM} by 1.9 to 3.2 points, more than the 1.3 points separating the families.
No family wins everywhere, but only the full mix beats \textsc{OpenTSLM} on all eight binary tasks, and cross-channel captions, which \textsc{TimeCap} adds, match structural ones on average.
Labels alone stay ahead on the ordinal targets, and per-week differences are small and mixed.

\begin{table}[h!t]
    \scriptsize
    \centering
    \caption{\textbf{Caption-family ablation (1B, user level).} Per-user AUROC (binary) and Spearman~$\rho$ (ordinal) as in Table~\ref{tab:opentslm_comparison_user}, whose 1B \textsc{OpenTSLM-Cap} is re-evaluated here as \textit{All}.
    One seed per arm; best per row in bold. $^{\dagger}$Held-out task.}
    \vspace{-0.5em}
    \setlength{\tabcolsep}{4pt}
    \renewcommand{\arraystretch}{1.05}
    \begin{tabular*}{\textwidth}{@{\extracolsep{\fill}}lccccc@{}}
    \toprule
        & & \multicolumn{4}{c}{\textsc{OpenTSLM-Cap}, caption family} \\
        \cmidrule(lr){3-6}
    \textbf{Task} & \textsc{OpenTSLM} & Structural & Semantic & Cross-channel & All \\
    \midrule
    \multicolumn{6}{l}{\textit{Binary (AUROC)}} \\
    Biological Sex & 78.90 & 78.87 & 78.42 & 78.56 & \textbf{79.10} \\
    Employment & 64.74 & \textbf{69.09} & 66.61 & 66.30 & 66.33 \\
    Hypertension & 67.96 & 68.77 & \textbf{71.70} & 70.38 & 69.74 \\
    Sleep Diagnosis & 58.86 & 60.44 & 59.25 & \textbf{61.26} & 59.20 \\
    Cardiovascular Disease$^{\dagger}$ & 62.09 & \textbf{66.53} & 60.16 & 64.22 & 64.84 \\
    Coronary Artery Disease$^{\dagger}$ & 63.38 & 62.80 & 59.87 & 62.06 & \textbf{65.75} \\
    Atrial Fibrillation$^{\dagger}$ & 59.82 & 63.72 & 64.35 & \textbf{68.83} & 68.43 \\
    Diabetes$^{\dagger}$ & 57.19 & \textbf{68.02} & 67.70 & 66.41 & 64.16 \\
    \midrule
    Mean, all binary & 64.12 & \textbf{67.28} & 66.01 & 67.25 & 67.19 \\
    Mean, held-out$^{\dagger}$ & 60.62 & 65.27 & 63.02 & 65.38 & \textbf{65.79} \\
    \midrule
    \multicolumn{6}{l}{\textit{Ordinal (Spearman $\rho$)}} \\
    BMI (5 levels) & \textbf{62.30} & 59.47 & 60.89 & 61.85 & 60.56 \\
    Happiness (4 levels) & \textbf{13.30} & 8.08 & 10.41 & 10.44 & 9.13 \\
    \bottomrule
    \end{tabular*}
    \label{tab:caption_family_ablation}
    \vspace{-1em}
\end{table}

\FloatBarrier

\section{Discussion}
\label{sec:discussion}

With \textsc{TimeCap}, we contribute the first open and deterministic captioning pipeline that extends existing template hierarchies with a cross-channel synthesis tier and a channel-schema abstraction.
It offers a grounded, schema-driven alternative to LLM-driven pipelines, whose outputs cannot be reproduced, and thereby addresses both the introduced methods gap and the extensibility gap.
The practical value of \textsc{TimeCap} is evident in \textsc{MHCap}, an open paired-caption layer for multimodal consumer wearables comprising over seven million captions across 908{,}944 person-days from 2{,}081 participants and covering 19 HealthKit channels.
Building on this corpus, caption-augmented training of \textsc{OpenTSLM}~\citep{openTSLM} outperforms an architecturally identical label-only baseline on 29 of 32 user-level binary evaluations (\S\ref{sec:metabolic_task}), with the largest gains on zero-shot clinical tasks.
In other words, the captions help the model understand sensor data, which also benefits tasks it was never trained on.

Ultimately, our work makes research on sensor-language models more accessible as well as transparent.
First, \textsc{MHCap} supports training and evaluating \glspl{TSLM} on real consumer wearable data in a research area that has so far depended on closed corpora.
Second, since \textsc{TimeCap} is deterministic, every caption can be regenerated and traced back to the signal, which makes training data for health models auditable.
Third, a corpus whose captions improve generalization to unseen clinical conditions lowers the barrier for studying new health outcomes at the population level in both research and downstream applications.

\paragraph{Limitations}
Our quantitative comparison leaves two adjacent baselines untested, namely LLM-generated captions of the same signals and frontier \glspl{VLM} prompted on rendered signal-plot images, which would quantify the determinism-versus-lexical-diversity trade-off we deliberately make.
Captions also come from fixed template banks with 20 variants per caption type, which limits their linguistic diversity.
In additional zero-shot experiments not reported in Section~\ref{sec:results}, ordinal transfer to held-out blood-pressure and sleep-duration categories furthermore failed across all configurations, with parse rates collapsing and predictions defaulting to the majority class, indicating that caption supervision alone is not yet sufficient for clinical labels with strong class imbalance.

Furthermore, the \gls{MHC} cohort is skewed toward white, male participants from the US with a median age of 39 years
(Appendix~\ref{app:cohort}), and models pretrained on the corpus may not generalize uniformly to underrepresented populations.
In addition, a per-user AUROC between 59.4 and 70.5 on the clinical tasks (Hypertension, Sleep Diagnosis, and the four held-out conditions) indicates population-level patterns rather than reliable individual predictions.
Any individual-level use would therefore require independent prospective validation and safeguards against misuse, for example by insurers or employers (Appendix~\ref{app:responsible_use}).

Future work could add a per-channel data-availability gate before structural detection, which would prevent missing or zero-filled periods from being captioned as events, and constrained language-model post-processing, which could make captions more readable while keeping them grounded in the signal.
Applying \textsc{TimeCap} to other cohorts and devices only requires new channel definitions, while high-frequency signals such as ECG would additionally need dedicated feature extraction.
Finally, the same pipeline could caption time series beyond health, such as the industrial and energy data mentioned in the introduction.

\section{Conclusion}
We close the data, methods, and extensibility gaps for paired wearable-sensor captions with \textsc{TimeCap}, a deterministic captioning pipeline with no language model in the generation loop, and \textsc{MHCap}, an open paired-caption layer for multimodal consumer wearables, further demonstrated on \gls{WESAD} as a concrete instantiation of the pipeline's schema-agnostic extensibility.
Caption-augmented \textsc{OpenTSLM} outperforms its label-only baseline on 29 of 32 user-level binary evaluations, with held-out clinical gains suggesting that the captions help the model understand sensor data beyond the tasks it was trained on.
The pipeline provides a reproducible mechanism by which the community can regenerate, audit, and extend paired-caption resources alongside the next generation of wearable hardware.

\section*{Acknowledgments}
These data were contributed by users of the My Heart Counts Mobile Health Application study developed by Stanford University.
The manuscript was prepared using the \emph{A Reproducible Workflow for Evidence-Centered Scientific Papers}~\citep{schmiedmayer2026papertemplate} template.

\bibliographystyle{\paperbibliographystyle}
\bibliography{references}

\clearpage
\appendix

\section{Training}
\label{app:training_details}

Both \textsc{OpenTSLM} and \textsc{OpenTSLM-Cap} are fine-tuned for one epoch on their respective training mixes (label-only vs.\ caption + label), with all updates routed through LoRA adapters~\citep{hu2021lora} ($r=16$, $\alpha=32$).
Optimization uses AdamW with a linear warmup-linear decay schedule (warmup over the first $3\%$ of total steps), gradient-clip norm $1.0$, weight decay $10^{-2}$, batch size $8$, and per-group learning rates of $2{\times}10^{-4}$ for the time-series encoder and $1{\times}10^{-4}$ for the projector and LoRA parameters.
Each configuration was trained on a single NVIDIA H100 80\,GB for approximately 35 hours, totaling roughly 140 GPU-hours across the four reported (model, scale) configurations.
The two runs at each scale are identical in every other respect, isolating the effect of the caption supervision.
Each task is posed to the model as a natural-language question with a fixed set of admissible answer strings (e.g., \textit{``Is this person biologically male or female?''} with answers \textit{Male} / \textit{Female}); the per-task question templates and admissible-answer mappings are shown in Figure~\ref{fig:tslm-prompt}.

\begin{figure}[!htbp]
\centering
\fbox{\begin{minipage}{0.95\linewidth}
\textbf{\textsc{OpenTSLM} Prompt}\par\medskip
\hrule\medskip
\small
The model is shown one week (168 hours) of multi-channel wearable sensor data interleaved with text.
The prompt has three parts.

\textbf{System pre-prompt.}
\begin{quote}\small\ttfamily
You are an expert in wearable sensor time series analysis.
\end{quote}

\textbf{Per-channel time-series prompts.} One block per channel, with the soft-prompted time-series tokens appended after each text segment:
\begin{itemize}\itemsep0pt
    \item \emph{Continuous channels} (iPhone step count, iPhone distance, flights climbed, Apple Watch step count, Apple Watch distance, heart rate, active energy):\\
    \texttt{The following is the $\langle$channel$\rangle$ in $\langle$unit$\rangle$, it has mean $\langle\mu\rangle$ and std $\langle\sigma\rangle$:}
    \item \emph{Binary channels} (Asleep, InBed, Walking, Cycling, Running, Other, Mixed Cardio, Strength, Elliptical, HIIT, Functional, Yoga):\\
    \texttt{The following is a per-hour binary mask for $\langle$channel$\rangle$ (1 = active, 0 = inactive or nonwear):}
\end{itemize}

\textbf{Task post-prompt.} Task-specific question followed by the admissible answer set:
\begin{itemize}\itemsep2pt
    \item \emph{Binary (BiologicalSex):}\\
    \texttt{Is this person biologically male or female?\\Possible answers: Male, Female}
    \item \emph{Ordinal (BMI\_categories):}\\
    \texttt{Which BMI category best matches this participant: Underweight, Normal weight, Overweight, Obesity, or Morbid Obesity?\\Possible answers: Underweight, Normal weight, Overweight, Obesity, Morbid Obesity}
    \item \emph{Held-out clinical (Diabetes):}\\
    \texttt{Does this person have diabetes?\\Possible answers: True, False}
\end{itemize}

\textbf{Response rules} (appended to every post-prompt):
\begin{quote}\small\ttfamily
- Base your answer on the sensor patterns above.\\
- You MUST give a classification even if the signal is unclear - state limitations but still make your best guess.\\
- Never respond with a question.\\
- You MUST respond with ONLY <label>, exactly as written in the listed possible answers.
\end{quote}
\end{minipage}}
\caption{Prompt template used for \textsc{OpenTSLM} and \textsc{OpenTSLM-Cap} evaluation.}
\label{fig:tslm-prompt}
\end{figure}

\section{Evaluation Protocol}
\label{app:eval_protocol}

\subsection{Splits}
\label{app:splits}

We use participant-level train/validation/test splits at a 60/10/30 ratio with seed 42, restricted to participants who consented to data sharing with all qualified researchers~\cite{schuetz2026openmhcacceleratingsciencewearable}.
After applying \gls{MHC}-benchmark's data-quality filter (7-day windows containing at least 5 days that pass daily wear-time and per-channel variance thresholds; see Appendix~\ref{app:data_curation} and the benchmark protocol \citet{schuetz2026openmhcacceleratingsciencewearable}), 2{,}081 of the 11{,}894 sharable participants remain captionable, with 604 falling in the test partition; no participant appears in more than one split, eliminating subject-level leakage.
Per-task evaluation N varies further (279--587 users) due to label availability, see Table~\ref{tab:test_set_sizes_binary}.

\begin{table}[h!t]
    \centering
    \small
    \caption{\textbf{Per-task test-set composition: binary tasks.} Captioned test partition: 604 unique users; per-task counts vary due to per-label NaN filtering. ``n\_pos'' is the minority-class user count (the convention used for AUPRC reporting); the minority label is given in parentheses where it differs from the codebase default. $^{\dagger}$Held-out clinical task, not seen during fine-tuning.}
    \label{tab:test_set_sizes_binary}
    \begin{tabular}{lrrr}
    \toprule
    \textbf{Task} & \textbf{n\_users} & \textbf{n\_pos} & \textbf{prev. (\%)} \\
    \midrule
    Biological Sex (Female=pos)             & 587 & 149 & 25.4 \\
    Hypertension                            & 279 & 90  & 32.3 \\
    Employment (False=pos)                  & 531 & 127 & 23.9 \\
    Sleep Diagnosis                         & 530 & 101 & 19.1 \\
    Diabetes$^{\dagger}$                    & 279 & 23  & 8.2 \\
    Cardiovascular Disease$^{\dagger}$      & 464 & 71  & 15.3 \\
    Coronary Artery Disease$^{\dagger}$     & 464 & 32  & 6.9 \\
    Atrial Fibrillation$^{\dagger}$         & 464 & 15  & 3.2 \\
    \bottomrule
    \end{tabular}
\end{table}

\begin{table}[h!t]
    \centering
    \small
    \caption{\textbf{Per-task test-set composition: ordinal tasks.} User counts per ordinal level.
    Spearman~$\rho$ and QWK reliability scales with the spread of both predicted and ground-truth scores, so heavily concentrated tasks (e.g., Happiness, with 75\% of users in High/Very High) have lower effective signal than tasks where the distribution is more even (e.g., BMI).}
    \label{tab:test_set_sizes_ordinal}
    \begin{tabular}{lrp{0.62\textwidth}}
    \toprule
    \textbf{Task} & \textbf{n\_users} & \textbf{Level distribution (n, \%)} \\
    \midrule
    BMI (5 levels)       & 557 & Underweight 25\,(4.5\%); Normal weight 175\,(31.4\%); Overweight 206\,(37.0\%); Obesity 121\,(21.7\%); Morbid Obesity 30\,(5.4\%) \\
    Happiness (4 levels) & 328 & Low 28\,(8.5\%); Medium 53\,(16.2\%); High 133\,(40.5\%); Very High 114\,(34.8\%) \\
    \bottomrule
    \end{tabular}
\end{table}

\subsection{Aggregation Conventions}
\label{app:aggregation}

\gls{TSLM} predictions are emitted at the (user, week) level.
Table~\ref{tab:opentslm_comparison} reports per-week metrics (F1, accuracy, Spearman~$\rho$, QWK) computed across all weekly predictions in the test split.
Table~\ref{tab:opentslm_comparison_user} reports user-level metrics: per-user AUROC and AUPRC are computed by aggregating each user's weekly binary predictions to a mean and ranking across users; per-user Spearman~$\rho$ uses the user's mean predicted class against the user's static label; per-user QWK is computed against the rounded user-mean.

\subsection{Data Curation}
\label{app:data_curation}
\textsc{MHCap} inherits the data-quality filters of the \textsc{OpenMHC} benchmark~\citep{schuetz2026openmhcacceleratingsciencewearable}.
A period counts as non-wear when all seven continuous channels are zero or missing for more than 30 consecutive minutes, and a day is retained only if it has at least 12 hours of estimated wear time and passes per-channel variance thresholds that remove near-constant sensor artifacts.
A week is captioned only if at least five of its seven days are retained, which leaves 2{,}081 of the 11{,}894 sharable participants.
On retained days, zero heart-rate values are set to missing.
Step, distance, and active-energy channels are set to missing for a whole day only if they are zero throughout that day, and zero sleep minutes are set to missing on days with less than three hours of detected sleep.
No values are imputed, so missing values remain explicit in the signal matrix (Section~\ref{sec:TimeCap_and_Corpora}).
Per-channel data availability can be recovered from the missingness mask of the corresponding \textsc{OpenMHC} week.
Zero-filled periods within a retained day can still be captioned as events, which motivates the per-channel data-availability gate discussed in Section~\ref{sec:discussion}.

\section{Responsible Use and Release}
\label{app:responsible_use}

\textsc{MHCap} is intended for research on aligning wearable time series with natural language, such as pretraining and evaluating \glspl{TSLM}.
All health outcomes in our evaluation come from the \textsc{OpenMHC} label set and are self-reported through in-app surveys and not derived by clinical examination or diagnostic testing~\citep{schuetz2026openmhcacceleratingsciencewearable}.
The tasks characterize self-reported participant attributes and do not forecast future clinical events.
Neither \textsc{MHCap} nor models trained on it are intended for diagnosis or for decisions about individuals.

\subsection{Cohort Composition}
\label{app:cohort}
The \textsc{OpenMHC} cohort reflects the population that chooses to use a digital health app and is not representative of the general public~\citep{schuetz2026openmhcacceleratingsciencewearable}.
The median participant is 39 years old.
Of the participants who reported their biological sex, 77.1\% are male, and of the 31.7\% who reported their ethnicity, 82.4\% are white.
About 90\% of participants are from the US, about 10\% from the UK, and less than 1\% from Hong Kong.
Half of the participants contributed less than one week of data, and 13.1\% contributed at least one year.
These figures describe all 11{,}894 \textsc{OpenMHC} participants.
The 2{,}081 participants in \textsc{MHCap} passed an additional data-quality filter (Appendix~\ref{app:data_curation}), so their composition may differ.
In our test split, 25.4\% of users with a biological-sex label are female (Table~\ref{tab:test_set_sizes_binary}), close to the 22.9\% in the full cohort.

\subsection{Data Availability and Access}
\label{app:data_access}
The underlying \textsc{OpenMHC} data are planned for release to qualified researchers under a research use agreement with the Stanford University School of Medicine~\citep{schuetz2026openmhcacceleratingsciencewearable}.
The \textsc{TimeCap} pipeline is publicly available under the MIT license.\footnote{\url{https://github.com/SchmiedmayerLab/TimeCAP}}
Precomputed \textsc{MHCap} captions will be released alongside \textsc{OpenMHC} under the same research-use terms.
Researchers with authorized access to the underlying data can regenerate the captions using the public pipeline.
These terms restrict use to personal, non-commercial research and prohibit re-identification as well as redistribution without prior written permission from the Stanford University School of Medicine.
They allow training generative models only if no data can be obtained or recreated from them, and they exclude use in diagnosis or patient care.
These safeguards rely on agreements and cannot technically prevent misuse.

\subsection{Privacy}
\textsc{MHCap} involves no new data collection.
Captions describe only the 19 sensor channels and contain no survey responses or demographic information.
Each caption record retains the participant identifier and date of the corresponding \textsc{OpenMHC} record (Section~\ref{sec:TimeCap_and_Corpora}).
In \textsc{OpenMHC}, dates are shifted by a random whole-week offset per participant, which prevents linking records to absolute calendar dates~\citep{schuetz2026openmhcacceleratingsciencewearable}.
Caption records follow the same date shift, and the caption text itself contains no calendar dates.

\section{Dataset-Specific Configurations}
\label{app:dataset_configs}

\subsection{\textsc{MHCap} Configuration}
\label{app:snsrcap_config}

The \gls{MHC} instantiation of \textsc{TimeCap} uses dataset-specific channel mappings, aggregation rules (Table~\ref{tab:snsr-aggregation-config}), structural-detector thresholds, and cross-channel synthesizers while preserving the same statistical, structural, semantic, and cross-channel pipeline described in Section~\ref{sec:TimeCap_and_Corpora}.
For readability, the configuration tables use compact channel aliases: \texttt{IP} denotes iPhone-derived channels and \texttt{W} denotes Apple Watch-derived channels.
The released implementation maps these aliases to the corresponding HealthKit identifiers.

\textcolor{statistical-orange-text}{\textbf{Statistical}} aggregation is configured per channel family.
In the current implementation, continuous movement and energy channels use the default metric aggregator (mean, max, min, std), while heart rate uses non-zero aggregation to avoid bias from zero-valued entries.
Sleep/workout label channels are handled through semantic interval extraction and cross-channel synthesis (not via statistical aggregation).
Template banks contain multiple deterministic variants per annotation family, introducing controlled linguistic diversity while preserving the same underlying signal content.

\begin{table}[!htbp]
  \centering
  \small
  \caption{\textsc{OpenMHC} statistical aggregation configuration (current implementation).}
  \label{tab:snsr-aggregation-config}
  \resizebox{\textwidth}{!}{\begin{tabular}{p{0.40\textwidth}p{0.28\textwidth}p{0.28\textwidth}}
    \toprule
    \textbf{Channels} & \textbf{Aggregator} & \textbf{Extractor output} \\
    \midrule
    IPSteps, IPDist, IPFlights, WSteps, WDist, WEnergy & \texttt{MetricAggregator} (default) & Mean, max, min, std \\
    WHR & \texttt{NonZeroAggregator} & Non-zero mean, max, min, std \\
    \bottomrule
  \end{tabular}}
\end{table}

\textcolor{structural-blue-text}{\textbf{Structural}} spike detection is configured only for continuous channels.
Table~\ref{tab:snsr-spike-config} reports detector parameters, while the rationale for each threshold is given below.

\begin{table}[!htbp]
  \centering
  \small
  \caption{\textsc{OpenMHC} spike-detector configuration.
  All spike detectors use \texttt{min\_width=1}, \texttt{top\_k=4}.}
  \label{tab:snsr-spike-config}
  \resizebox{\textwidth}{!}{\begin{tabular}{lllllll}
    \toprule
    \textbf{Channels} & \texttt{filter\_zeros} & \texttt{min\_height} & \texttt{min\_prominence} & \texttt{min\_distance} & \textbf{Unit} \\
    \midrule
    IPSteps, WSteps & False & 100.0 & 0.0 & 10 & steps/min \\
    IPDist, WDist   & False & 72.0  & 0.0 & 10 & m/min \\
    IPFlights       & False & 1.0   & 0.0 & 10 & count/min \\
    WHR             & True  & 100.0 & 5.0 & 20 & bpm \\
    WEnergy         & False & 3000.0 & 0.0 & 15 & cal/min \\
    \bottomrule
  \end{tabular}}
\end{table}

The \textbf{spike thresholds} are intended as conservative descriptive heuristics rather than clinical or diagnostic cutoffs.
For step count, \texttt{min\_height=100} steps/min follows the common cadence heuristic that approximately 100 steps/min corresponds to moderate-intensity walking in adults~\citep{Tudor-Locke2020}.
For distance, \texttt{min\_height=72} m/min corresponds to 1.2 m/s, which lies within typical preferred walking-speed ranges and provides a simple distance-based analog to the cadence threshold~\citep{ZIEGLER2024111976}.
For flights climbed, \texttt{min\_height=1} flight/min marks discrete stair-climbing events.
For heart rate, \texttt{min\_height=100} bpm is used as an elevated-heart-rate heuristic, with \texttt{min\_prominence=5} bpm to suppress minor fluctuations around the threshold~\citep{page2016acchrs}.
For active energy, \texttt{min\_height=3000} cal/min corresponds to 180 kcal/hour above rest.
Since active energy excludes resting expenditure, this is a conservative threshold relative to a gross MET-based moderate-intensity benchmark: for a 60--70 kg individual, 3 METs corresponds to approximately 180--210 kcal/hour in gross expenditure, but only about 120--140 kcal/hour above rest.
The threshold therefore prioritizes periods with clearly elevated active energy expenditure~\citep{Herrmann2024-qk}.

\textbf{Trend} detection uses the same scanning defaults across movement and physiology channels, with heart rate zero-filtered.
To keep the table compact, Table~\ref{tab:snsr-trend-config} reports only channel groups and effective parameter settings.

\begin{table}[!htbp]
  \centering
  \small
  \caption{\textsc{OpenMHC} trend-detector configuration.
  Unless overridden, trend detectors use \texttt{window\_fracs=(0.1, 0.25, 1.0)}, \texttt{min\_window=12}, and \texttt{stride\_frac=0.25}.}
  \label{tab:snsr-trend-config}
  \resizebox{\textwidth}{!}{\begin{tabular}{lllllll}
    \toprule
    \textbf{Channels} & \texttt{filter\_zeros} & \texttt{min\_coverage} & \texttt{min\_effect} & \texttt{min\_r2} & \textbf{Notes} \\
    \midrule
    IPSteps, IPDist, IPFlights, WSteps, WDist, WEnergy & False & 0.75 & 1.25 & 0.50 & Default \\
    WHR & True & 0.75 & 1.25 & 0.50 & Zero-filtered \\
    \bottomrule
  \end{tabular}}
  \\[0.3em]
  \parbox{\textwidth}{\footnotesize All listed trend detectors use \texttt{min\_span=10} and \texttt{merge\_gap=5}.}
\end{table}

\textcolor{semantical-purple-text}{\textbf{Semantic}} extractors operate on binary sleep and workout channels by identifying contiguous active intervals and filtering out intervals shorter than the configured minimum duration.
Cross-channel synthesizers then enrich these semantic intervals with temporally aligned evidence from supporting continuous channels.
Table~\ref{tab:snsr-cross-channel-synthesizers} summarizes the \textsc{OpenMHC} synthesizer definitions.

\begin{table}[!htbp]
\centering
\small
\caption{\textsc{OpenMHC} cross-channel synthesizer definitions.
All listed synthesizers are instantiated with \texttt{min\_duration=5} at minute resolution.}
\label{tab:snsr-cross-channel-synthesizers}
\begin{tabular}{p{0.22\textwidth}p{0.19\textwidth}p{0.22\textwidth}p{0.27\textwidth}}
\toprule
Synthesizer & Source channels & Supporting channels & Output labels \\
\midrule
\texttt{Sleep}
& SleepAsleep, SleepInBed
& WHR, WDist, WSteps
& \texttt{sleep\_asleep}, \texttt{sleep\_inbed}, \texttt{in\_bed\_not\_sleeping} \\
\texttt{Walking}
& Walk
& WHR, WDist, WSteps, WEnergy
& \texttt{walking} \\
\texttt{Cardio}
& Run, Cycle
& WHR, WDist, WEnergy; WSteps for Run
& \texttt{cardio\_running}, \texttt{cardio\_cycling} \\
\texttt{EnduranceActivity}
& Elliptical, MixedCardio
& WHR, WDist, WEnergy
& \texttt{endurance\_elliptical}, \texttt{endurance\_mixed\_cardio} \\
\texttt{StationaryActivity}
& HIIT, Strength, Functional
& WHR, WEnergy, WDist, WSteps
& \texttt{stationary\_hiit}, \texttt{stationary\_strength}, \texttt{stationary\_functional} \\
\texttt{MindBodyActivity}
& Yoga
& WHR, WEnergy, WDist, WSteps
& \texttt{mind\_body\_yoga} \\
\texttt{OtherActivity}
& Other
& WHR, WDist, WEnergy
& \texttt{other\_activity} \\
\bottomrule
\end{tabular}
\end{table}

\textcolor{synthesizer-green-text}{\textbf{Cross-channel synthesizers}} are grouped by the type of semantic interval they enrich.
Sleep synthesizers distinguish asleep, in-bed, and in-bed-not-asleep intervals, while activity synthesizers attach concurrent physiological and movement evidence to walking, cardio, endurance, stationary, mind-body, and other workout intervals.
This keeps the semantic event label grounded in the binary HealthKit channel while adding synchronized quantitative context from heart rate, distance, step count, and active energy where available.
\par

\subsection{\textsc{WESADCap} Configuration}
\label{app:wesadcap_config}

The WESAD instantiation of \textsc{TimeCap} uses dataset-specific channel mappings, per-channel aggregation rules, structural-detector thresholds, and state-conditioned cross-channel synthesizers while preserving the same statistical, structural, semantic, and cross-channel pipeline described in Section~\ref{sec:TimeCap_and_Corpora}.
In the released implementation, channels are represented by explicit WESAD identifiers and processed on a per-second timeline (\texttt{time\_unit=seconds}).

\textbf{Preprocessing and channel construction.}
Raw WESAD streams are transformed into second-level channels before caption extraction.
Continuous channels are binned to 1-second resolution using within-second mean aggregation, while state labels are mapped to one-hot binary channels by majority vote within each second.
Specifically:

\begin{itemize}
\item Chest streams (\texttt{ACC}, \texttt{EMG}, \texttt{EDA}, \texttt{Temp}, \texttt{Resp}) are resampled from the chest sampling grid (700 Hz) to per-second bins.
\item Wrist streams are resampled from their native rates to per-second bins: \texttt{ACC} (32 Hz), \texttt{BVP} (64 Hz), \texttt{EDA} (4 Hz), \texttt{TEMP} (4 Hz), and \texttt{HR} (1 Hz from E4 files).
\item Derived channels are constructed before binning: acceleration magnitudes (\texttt{acc\_mag}) are Euclidean norms over xyz axes; respiration and BVP magnitudes are rectified (absolute value).
\item Wrist acceleration is converted to g units via the implementation constant \texttt{1/64}.
\item Label channels (\texttt{baseline}, \texttt{stress}, \texttt{amusement}, \texttt{meditation}) are generated as binary per-second state indicators via majority vote over the original label samples in each second.
\item HR alignment uses the inferred session start offset from E4 alignment; non-finite values are retained as missing (\texttt{NaN}) and ignored by downstream extractors.
\end{itemize}

\textcolor{statistical-orange-text}{\textbf{Statistical aggregation}} is configured separately from structural detection.
Continuous channels use the default metric aggregation (mean, max, min, std), while wrist heart rate uses non-zero aggregation to avoid bias from zero-valued entries.
Label channels are primarily consumed by semantic and cross-channel extractors rather than statistical summaries.

\begin{table}[!htbp]
  \centering
  \small
  \caption{WESAD statistical aggregation configuration (current implementation).}
  \label{tab:wesad-aggregation-config}
  \begin{tabular}{lll}
    \toprule
    \textbf{Channels} & \textbf{Aggregator} & \textbf{Extractor output} \\
    \midrule
    \texttt{wesad:e4:hr} & \texttt{NonZeroAggregator} & Non-zero mean, max, min, std \\
    All other channels   & \texttt{MetricAggregator} (default) & Mean, max, min, std \\
    \bottomrule
  \end{tabular}
\end{table}

\textcolor{structural-blue-text}{\textbf{Structural detectors}} are configured only for continuous channels.
To keep the layout compact, trend and spike parameters are grouped by channel family.

\begin{table}[!htbp]
  \centering
  \small
  \caption{WESAD trend-detector configuration by channel group.}
  \label{tab:wesad-trend-config}
  \resizebox{\textwidth}{!}{\begin{tabular}{llllll}
    \toprule
    \textbf{Channels} & \texttt{filter\_zeros} & \texttt{min/max\_window} & \texttt{min\_effect} & \texttt{min\_r2} & \textbf{Notes} \\
    \midrule
    Chest/Wrist ACC axes and ACC magnitude & False & 40 / 150 & 1.8  & 0.40 & \texttt{min\_span=30}, \texttt{merge\_gap=15} \\
    Chest EMG                              & False & 20 / 240 & 0.6  & 0.15 & \texttt{min\_span=14}, \texttt{merge\_gap=15} \\
    Wrist HR (\texttt{wesad:e4:hr})        & True  & defaults & 1.25 & 0.50 & defaults: \texttt{min\_span=10}, \texttt{merge\_gap=5} \\
    Chest/Wrist EDA                        & False & 30 / 240 & 0.10 & 0.60 & \texttt{min\_span=40}, \texttt{merge\_gap=15} \\
    Chest/Wrist temperature                & False & 90 / 600 & 0.18 & 0.70 & \texttt{min\_span=60}, \texttt{merge\_gap=30} \\
    Chest respiration                      & False & 24 / 180 & 0.75 & 0.20 & \texttt{min\_span=16}, \texttt{merge\_gap=15} \\
    Wrist BVP                              & False & 20 / 300 & 0.7  & 0.15 & \texttt{min\_span=14}, \texttt{merge\_gap=15} \\
    \bottomrule
  \end{tabular}}
  \\[0.3em]
  \parbox{\textwidth}{\footnotesize Trend defaults (when not overridden): \texttt{window\_fracs=(0.1,0.25,1.0)}, \texttt{min\_window=12}, \texttt{stride\_frac=0.25}, \texttt{min\_coverage=0.75}, \texttt{min\_effect=1.25}, \texttt{min\_r2=0.5}, \texttt{min\_span=10}, \texttt{merge\_gap=5}.}
\end{table}

\begin{table}[!htbp]
  \centering
  \small
  \caption{WESAD spike-detector configuration.
  All spike detectors use \texttt{min\_width=1}, \texttt{top\_k=4}, and \texttt{min\_distance=60}.}
  \label{tab:wesad-spike-config}
  \begin{tabular}{p{0.28\textwidth}p{0.14\textwidth}p{0.14\textwidth}p{0.18\textwidth}p{0.08\textwidth}}
    \toprule
    Channels & \texttt{filter\_zeros} & \texttt{min\_height} & \texttt{min\_prominence} & Unit \\
    \midrule
    Chest ACC magnitude     & False & $-$ & 0.05  & g \\
    Wrist ACC magnitude     & False & $-$ & 0.10  & g \\
    Wrist HR                & True  & 100.0 & 5.0 & bpm \\
    Chest EDA               & False & $-$ & 0.25  & $\mu$S \\
    Wrist EDA               & False & $-$ & 0.50  & $\mu$S \\
    Chest/Wrist temperature & False & $-$ & 0.25  & $^\circ$C \\
    Chest respiration       & False & $-$ & 10.0  & a.u. \\
    Wrist BVP               & False & $-$ & 200.0 & a.u. \\
    \bottomrule
  \end{tabular}
\end{table}

The \textbf{spike thresholds} are intended as conservative descriptive heuristics rather than clinical or diagnostic cutoffs.
For acceleration magnitude, the chest and wrist spike prominences (\texttt{0.05}/\texttt{0.10} g) are set above the raw-accelerometry thresholds reported by ~\citep{Bakrania2016-ll}, who found wrist ENMO cut-points of approximately 26--53 mg for separating sedentary behavior from motion-based light-intensity activities; our thresholds therefore prioritize more salient movement excursions rather than all light movement events.
For HR, we reuse the same detector as in the \textsc{MHCap} configuration, with \texttt{min\_height=100} bpm and \texttt{min\_prominence=5} bpm, thereby preserving the same elevated-rate heuristic and citation rationale~\citep{page2016acchrs}.
For EDA, the chest and wrist channels appear to differ substantially in scale, so we use sensor-specific prominence thresholds (chest \texttt{0.25}, wrist \texttt{0.50}) to avoid narrating small scale-dependent fluctuations as salient events.
For temperature, \texttt{min\_prominence=0.25}\,\textdegree C is used to suppress small local fluctuations while still capturing salient thermal excursions, motivated by reported acute exercise-related skin-temperature changes that occur on the order of several tenths of a degree to about one degree Celsius depending on body site~\citep{CORRALPEREZ2023103678}.
For respiration and BVP, the signal values are expressed in arbitrary units; thresholds were therefore selected heuristically.
Higher prominence thresholds (\texttt{10.0} and \texttt{200.0}) help reduce false positives arising from oscillatory baselines and artifact-heavy waveform amplitudes.

\textbf{Trend} detection in WESAD follows a shared scanning framework, but channel-specific thresholds are used for both movement and physiology signals; heart rate is additionally zero-filtered.
To keep the table compact, Table~\ref{tab:wesad-trend-config} reports channel groups and their effective parameter settings.

\textcolor{semantical-purple-text}{\textbf{Semantic extractors}} operate on binary state-label channels by identifying contiguous active intervals and filtering out intervals shorter than the configured minimum duration.
Cross-channel synthesizers then enrich these state intervals with temporally aligned evidence from physiology and movement channels.

\begin{table}[!htbp]
  \centering
  \small
  \caption{WESAD cross-channel synthesizer definitions.
  All listed synthesizers are instantiated with \texttt{min\_duration=30} at second resolution.}
  \label{tab:wesad-cross-channel-synthesizers}
  \resizebox{\textwidth}{!}{\begin{tabular}{llp{0.35\textwidth}l}
    \toprule
    \textbf{Synthesizer} & \textbf{Source channel} & \textbf{Supporting channels} & \textbf{Output label} \\
    \midrule
    \texttt{StressResponseSynthesizer}         & \texttt{wesad:label:stress}    & HR, chest/wrist EDA, chest respiration, chest/wrist ACC magnitude & \texttt{stress\_response} \\
    \addlinespace
    \texttt{RecoveryCalmSynthesizer} (baseline)  & \texttt{wesad:label:baseline}  & HR, chest/wrist EDA, chest respiration, chest/wrist ACC magnitude & \texttt{calm\_baseline} \\
    \addlinespace
    \texttt{RecoveryCalmSynthesizer} (meditation) & \texttt{wesad:label:meditation} & HR, chest/wrist EDA, chest respiration, chest/wrist ACC magnitude & \texttt{calm\_meditation} \\
    \addlinespace
    \texttt{AmusementActivationSynthesizer}    & \texttt{wesad:label:amusement} & HR, chest/wrist EDA, chest/wrist ACC magnitude & \texttt{amusement\_activation} \\
    \bottomrule
  \end{tabular}}
\end{table}

\textcolor{synthesizer-green-text}{\textbf{Cross-channel synthesizers}} are grouped by affective-state semantics: stress activation, calm recovery (baseline/meditation), and amusement activation.
Each synthesized interval remains grounded in the active binary state channel, while metric suffixes add synchronized context from HR, EDA, respiration, and movement, with chest-vs-wrist channel choice resolved adaptively from the strongest available evidence in each interval.
\par

\clearpage
\section{Generated Caption}
\label{app:generated_captions}

Time-aligned examples of captions generated by \textsc{TimeCap} for each dataset are presented below.

\subsection{\textsc{MHCap} Captions}
\label{app:snsrcap_captions}

\begin{figure}[!htbp]
    \centering
    \includegraphics[width=\linewidth,height=0.35\textheight,keepaspectratio,alt={Wearable sensor traces for sleep example 1 with highlighted intervals paired with statistical, structural, semantic, and cross-channel captions.}]{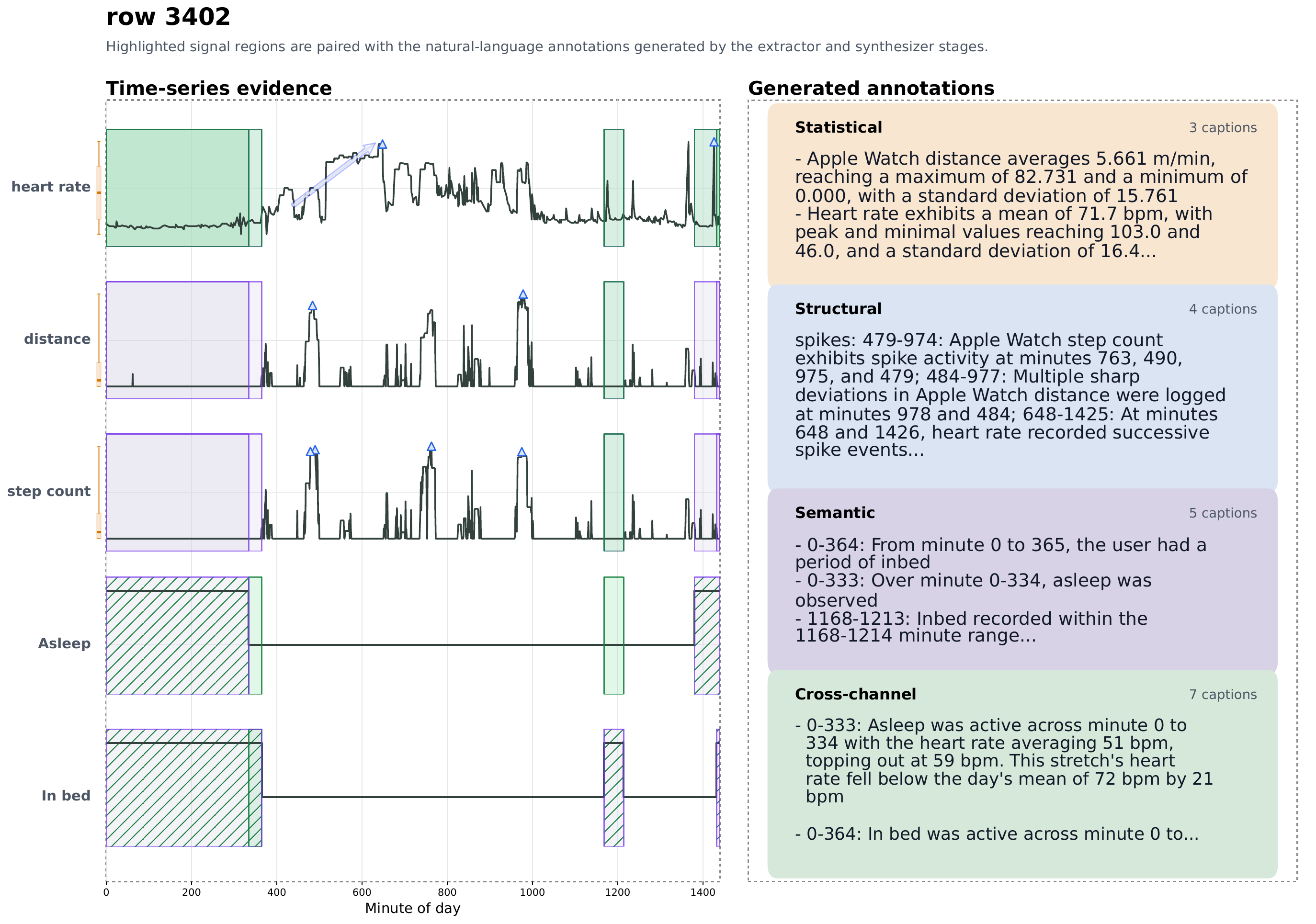}
    \caption{Example with sleep caption 1}
    \label{fig:SENSORCap1}
\end{figure}

\begin{figure}[!htbp]
    \centering
    \includegraphics[width=\linewidth,height=0.38\textheight,keepaspectratio,alt={Wearable sensor traces for walking example 1 with highlighted intervals paired with statistical, structural, semantic, and cross-channel captions.}]{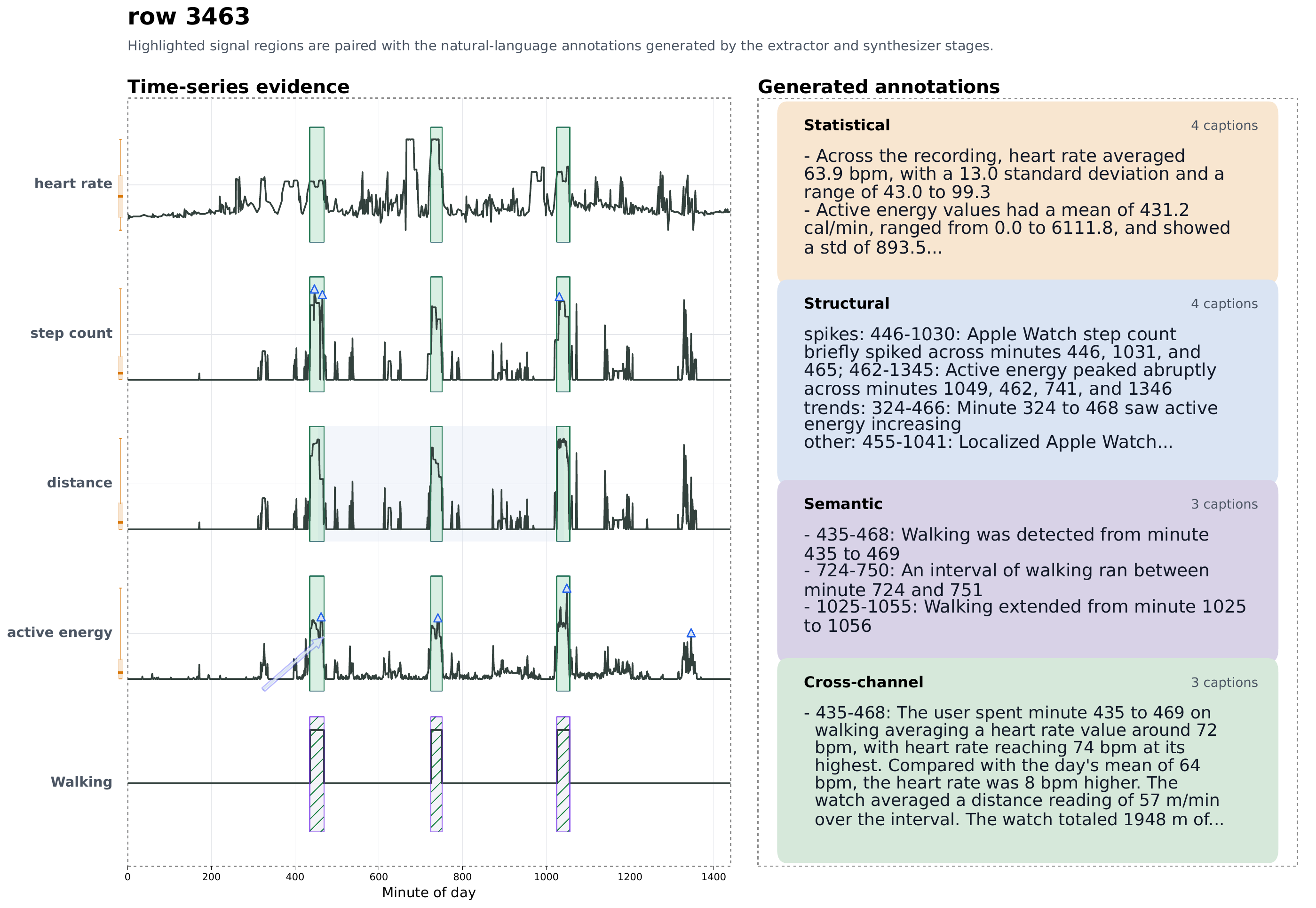}
    \caption{Example with walking 1}
    \label{fig:SENSORCap-walking1}
\end{figure}

\begin{figure}[!htbp]
    \centering
    \includegraphics[width=\linewidth,height=0.38\textheight,keepaspectratio,alt={Wearable sensor traces for yoga with highlighted intervals paired with statistical, structural, semantic, and cross-channel captions.}]{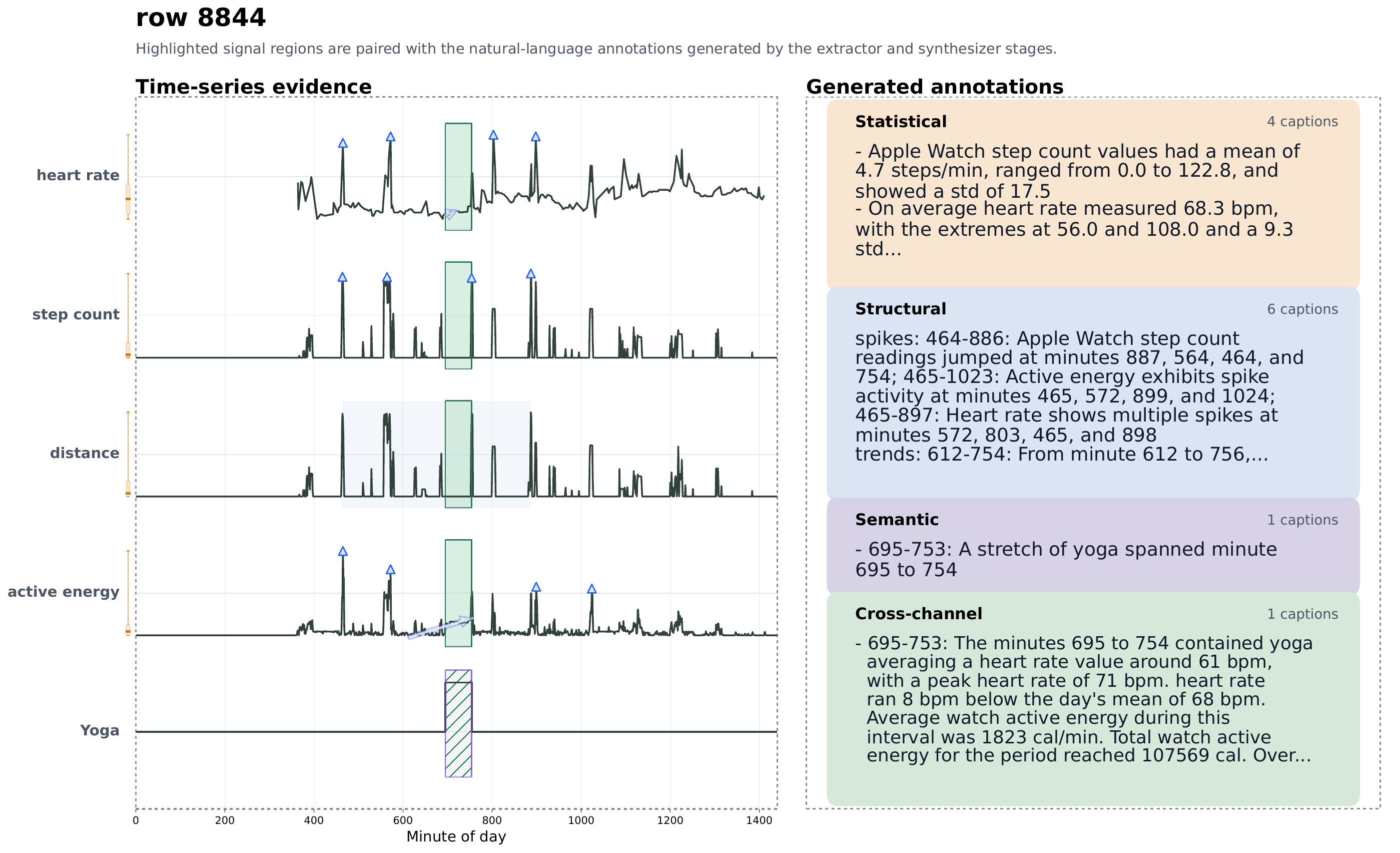}
    \caption{Example with yoga}
    \label{fig:SENSORCap10}
\end{figure}

The captioned examples for \textsc{MHCap} can be seen in Figures \ref{fig:SENSORCap1}-\ref{fig:SENSORCap10}.

\begin{figure}[!htbp]
    \centering
    \includegraphics[width=\linewidth,height=0.35\textheight,keepaspectratio,alt={Wearable sensor traces for sleep example 2 with highlighted intervals paired with statistical, structural, semantic, and cross-channel captions.}]{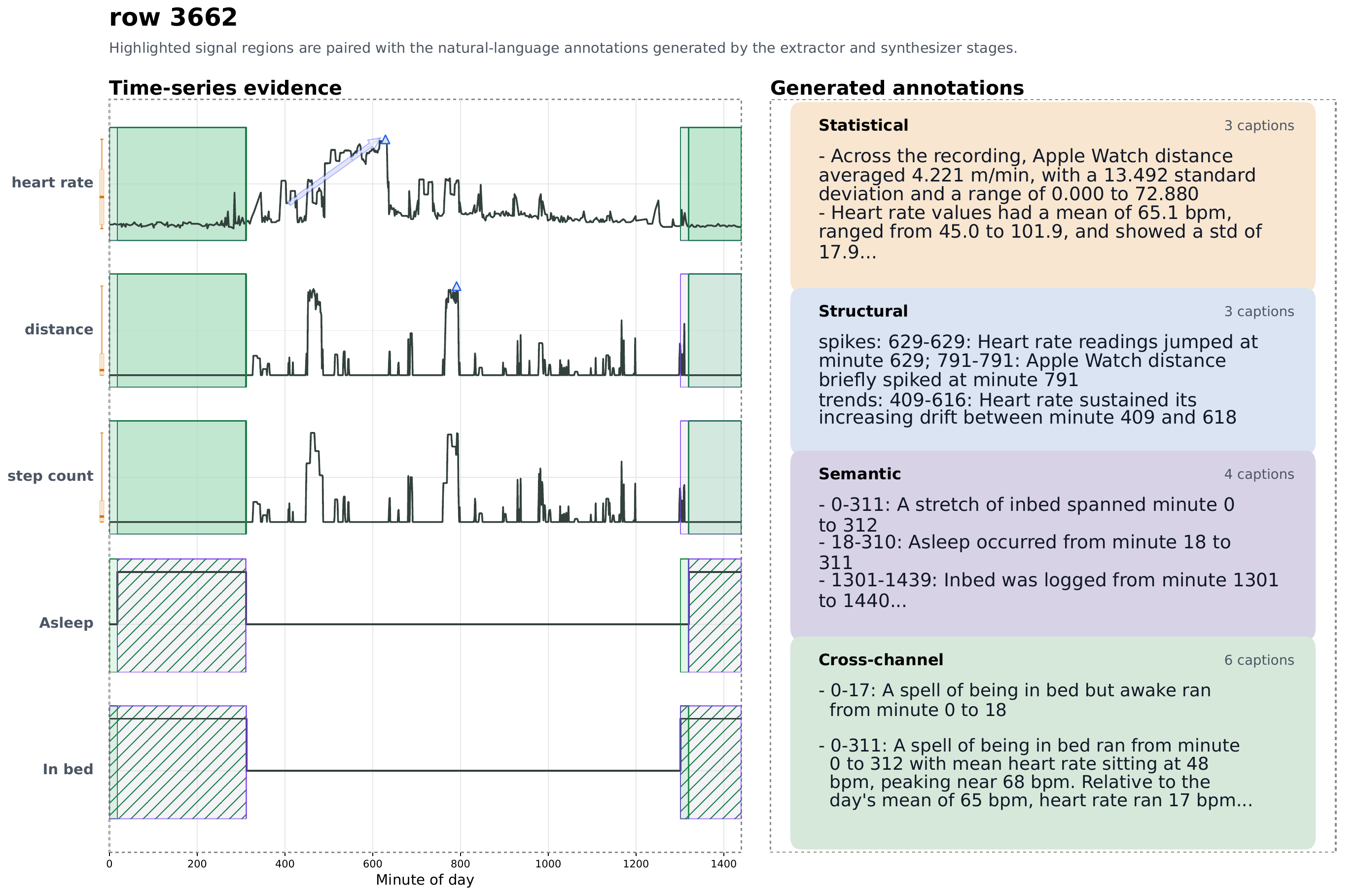}
    \caption{Example with sleep caption 2}
\end{figure}
\clearpage

\begin{figure}[!htbp]
    \centering
    \includegraphics[width=\linewidth,height=0.38\textheight,keepaspectratio,alt={Wearable sensor traces for sleep example 3 with highlighted intervals paired with statistical, structural, semantic, and cross-channel captions.}]{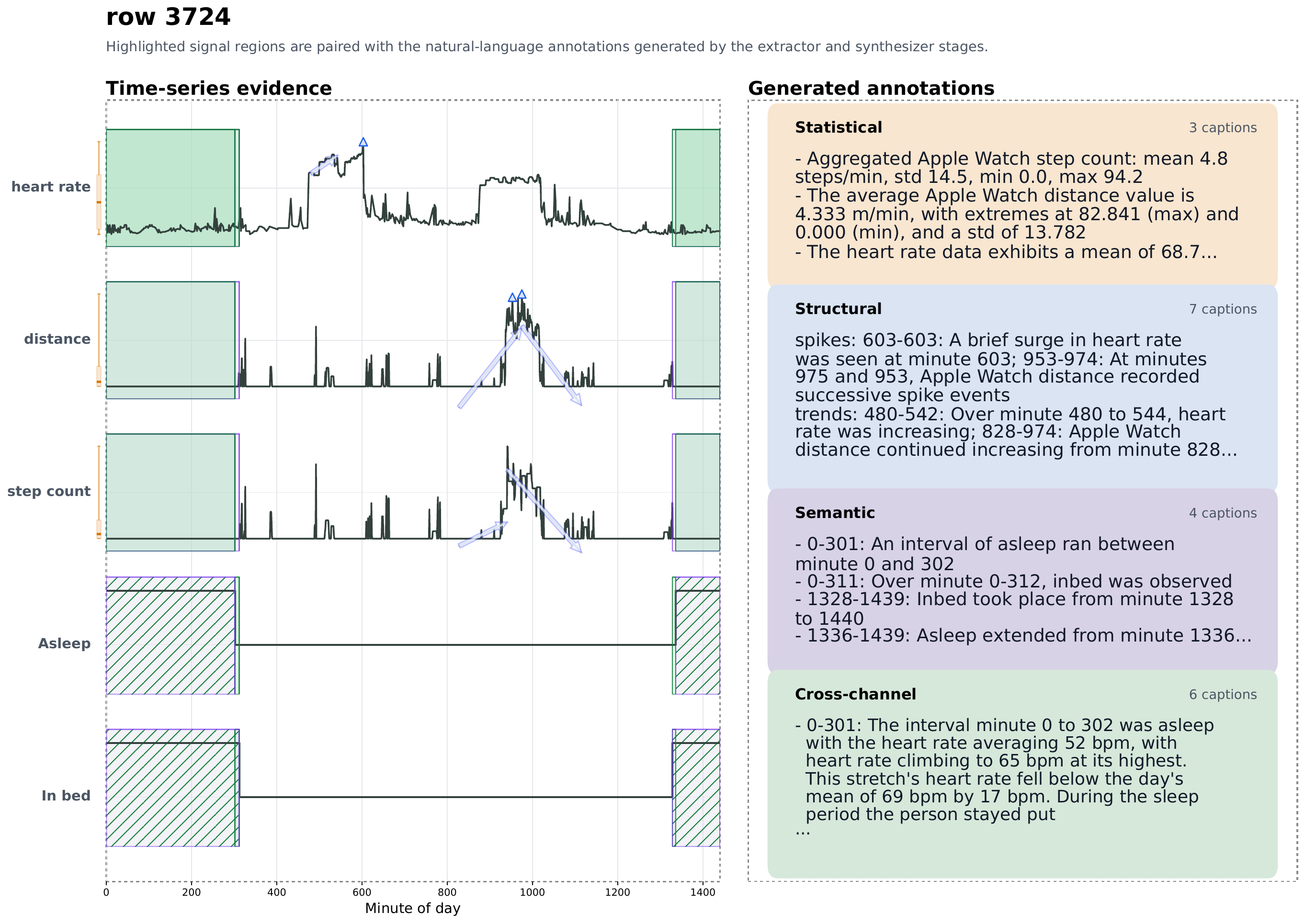}
    \caption{Example with sleep caption 3}
\end{figure}

\begin{figure}[!htbp]
    \centering
    \includegraphics[width=\linewidth,height=0.38\textheight,keepaspectratio,alt={Wearable sensor traces for cycling with highlighted intervals paired with statistical, structural, semantic, and cross-channel captions.}]{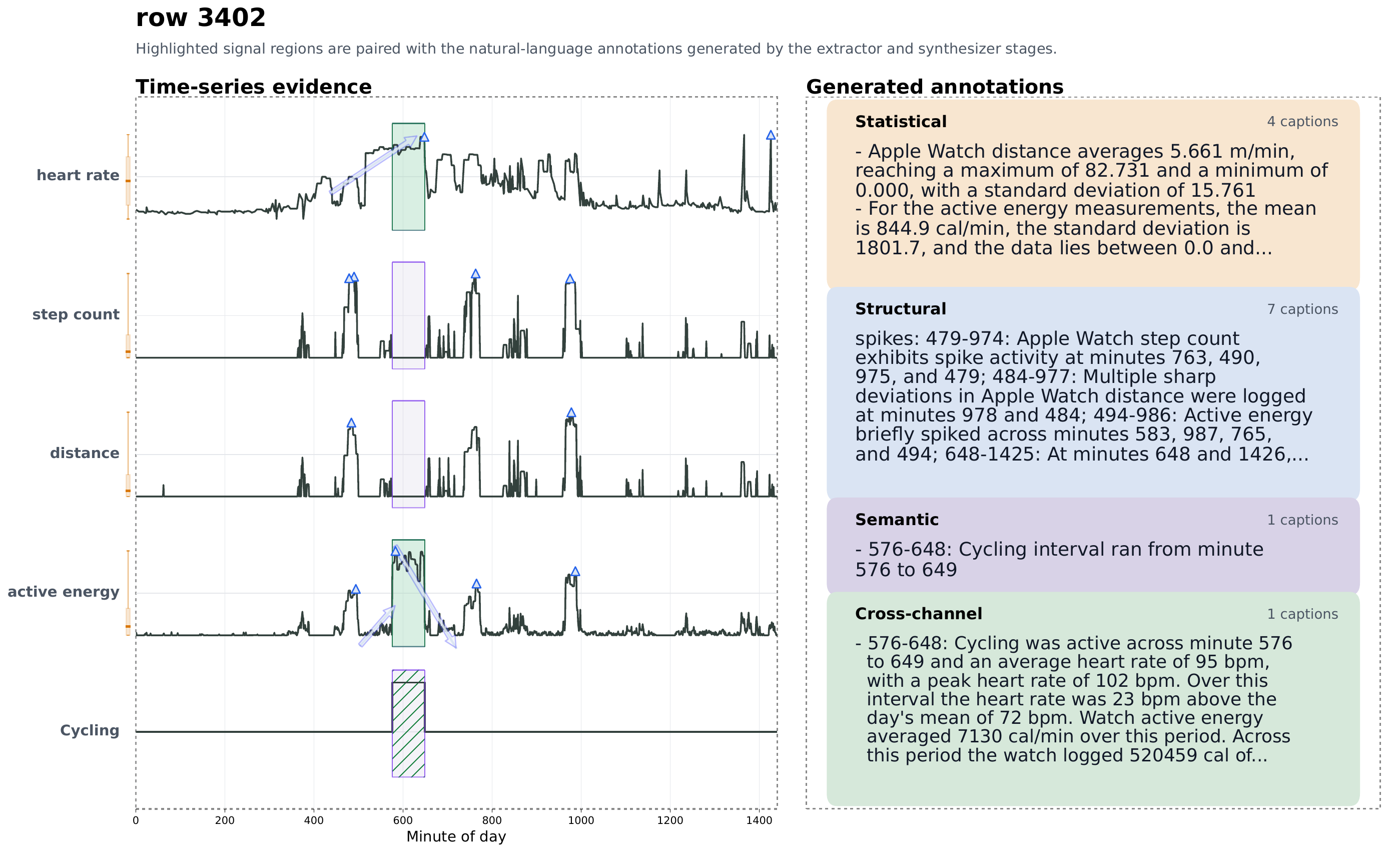}
    \caption{Example with cycling}
\end{figure}
\clearpage

\begin{figure}[!htbp]
    \centering
    \includegraphics[width=\linewidth,height=0.38\textheight,keepaspectratio,alt={Wearable sensor traces for elliptical exercise with highlighted intervals paired with statistical, structural, semantic, and cross-channel captions.}]{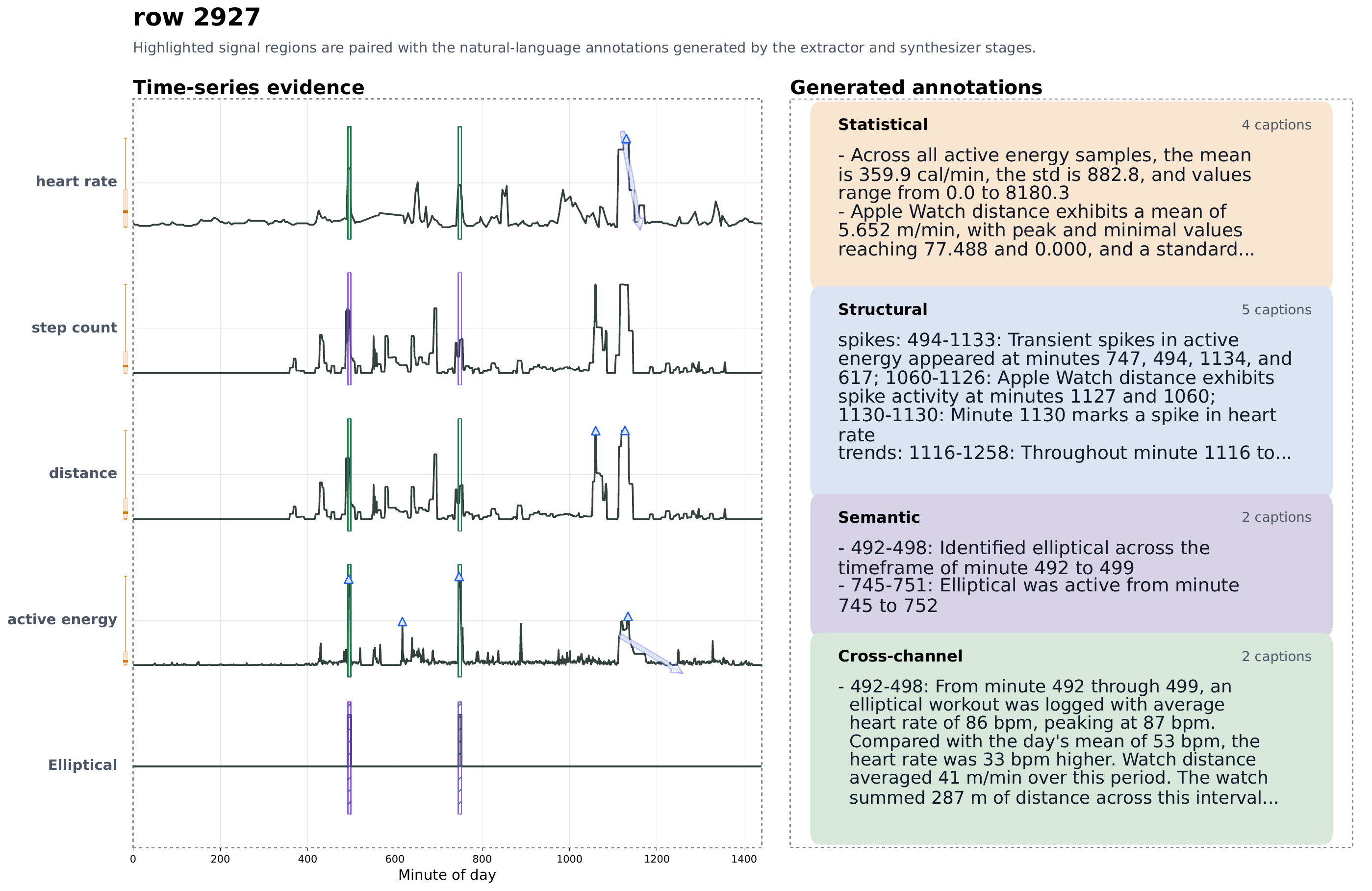}
    \caption{Example with elliptical}
\end{figure}

\begin{figure}[!htbp]
    \centering
    \includegraphics[width=\linewidth,height=0.38\textheight,keepaspectratio,alt={Wearable sensor traces for running with highlighted intervals paired with statistical, structural, semantic, and cross-channel captions.}]{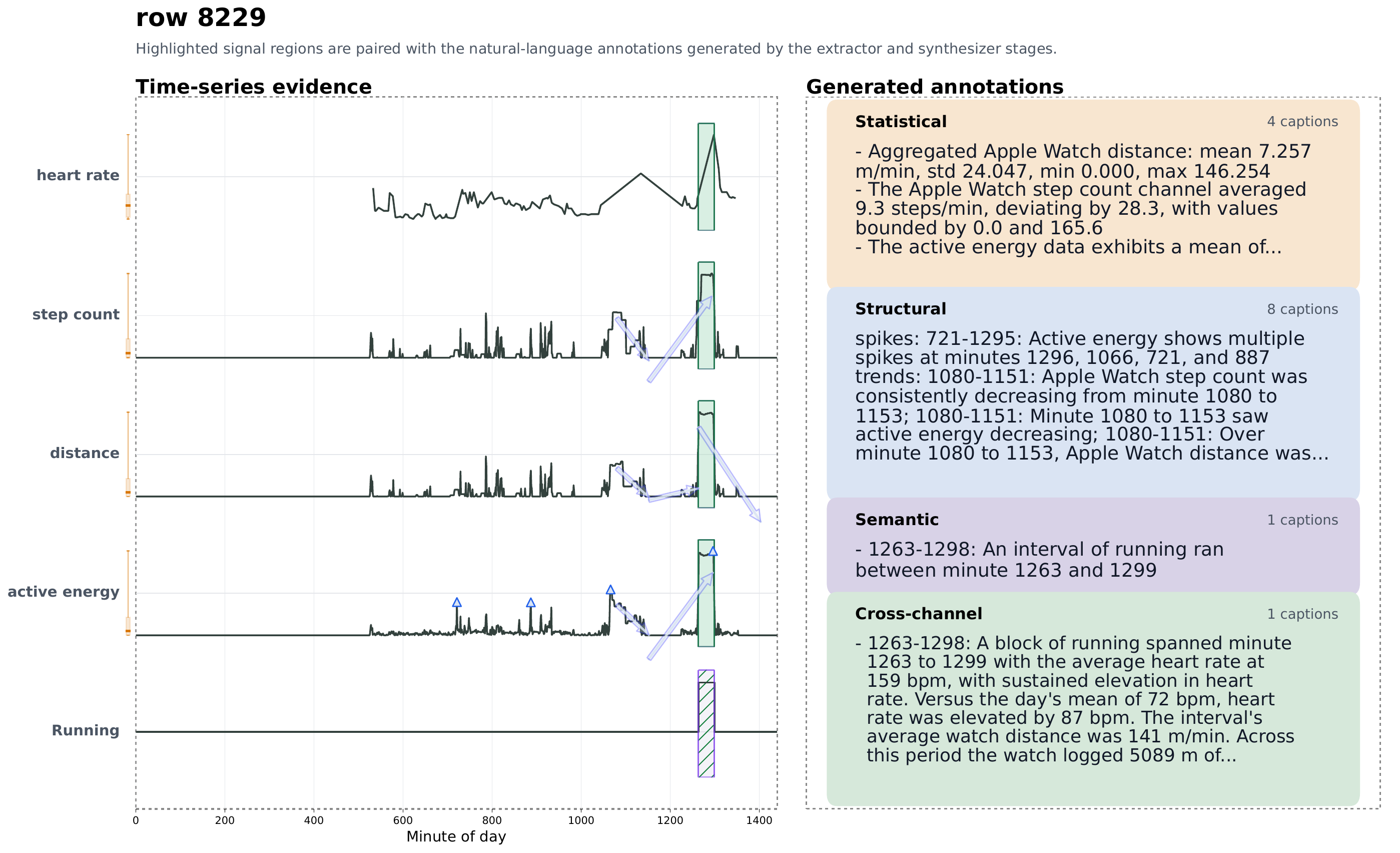}
    \caption{Example with running}
\end{figure}
\clearpage

\begin{figure}[!htbp]
    \centering
    \includegraphics[width=\linewidth,height=0.38\textheight,keepaspectratio,alt={Wearable sensor traces for high-intensity interval training with highlighted intervals paired with statistical, structural, semantic, and cross-channel captions.}]{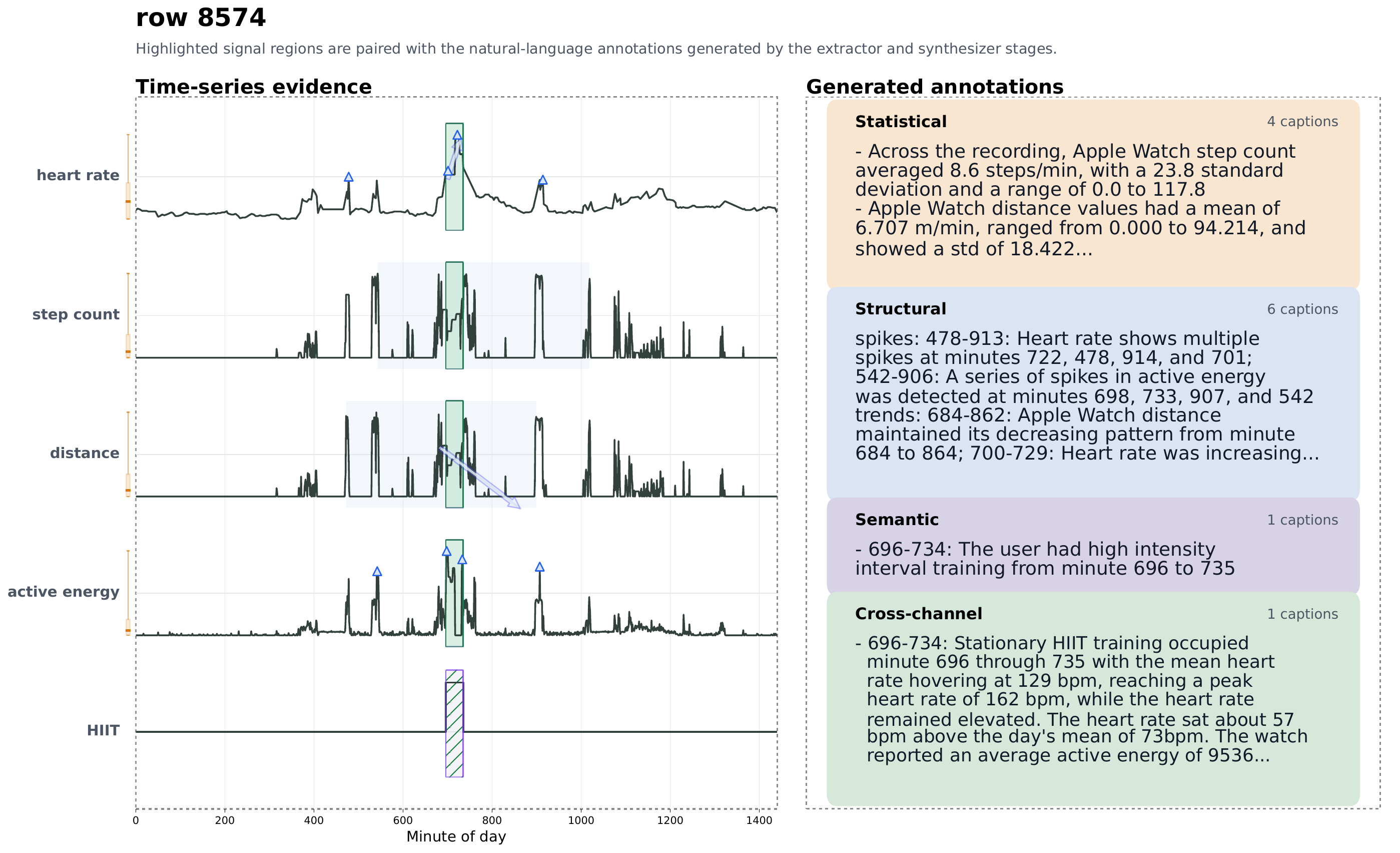}
    \caption{Example with HIIT}
\end{figure}

\clearpage

\begin{figure}[!htbp]
    \centering
    \includegraphics[width=\linewidth,height=0.38\textheight,keepaspectratio,alt={Wearable sensor traces for walking example 2 with highlighted intervals paired with statistical, structural, semantic, and cross-channel captions.}]{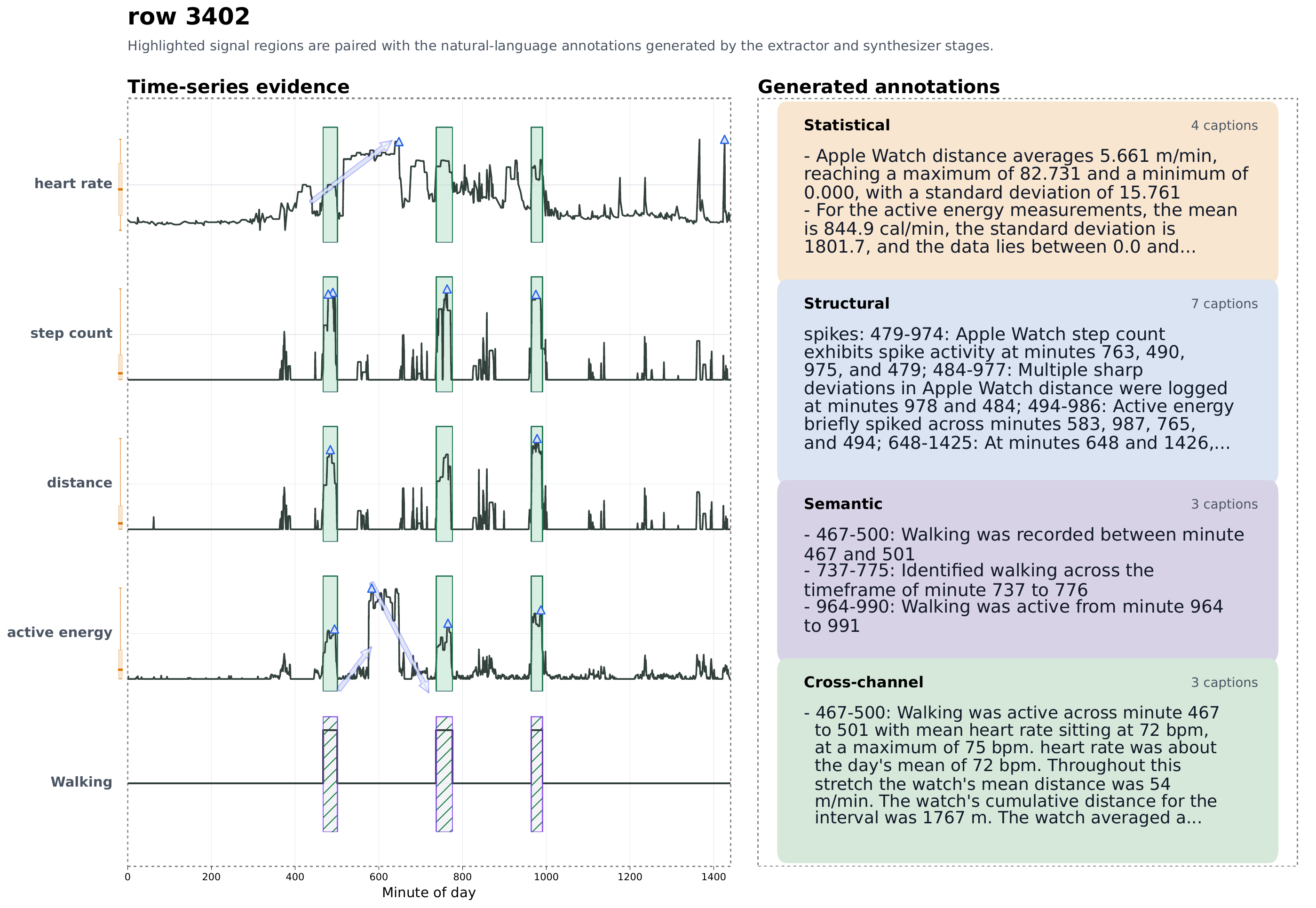}
    \caption{Example with walking 2}
\end{figure}

\FloatBarrier
\subsection{\textsc{WESADCap} Captions}
\label{app:WESADcap_captions}

\begin{figure}[!htbp]
    \centering
    \includegraphics[width=\linewidth,height=0.80\textheight,keepaspectratio,alt={WESAD subject 1 physiological and state traces with highlighted intervals paired with statistical, structural, semantic, and cross-channel captions.}]{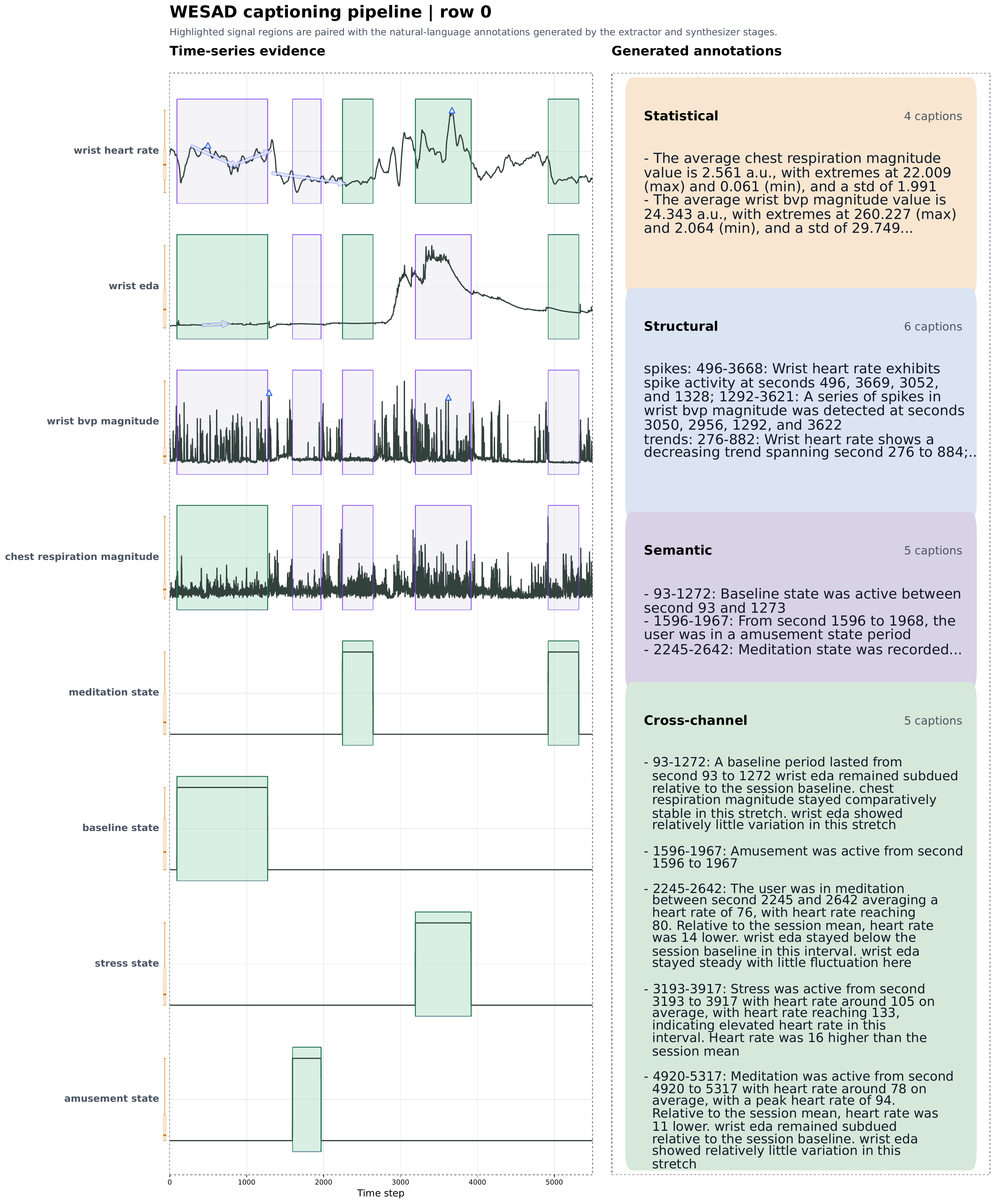}
    \caption{\textsc{WESADCap} subject 1.}
    \label{fig:WESADCap1}
\end{figure}

\begin{figure}[!htbp]
    \centering
    \includegraphics[width=\linewidth,alt={WESAD subject 15 physiological and state traces with highlighted intervals paired with statistical, structural, semantic, and cross-channel captions.}]{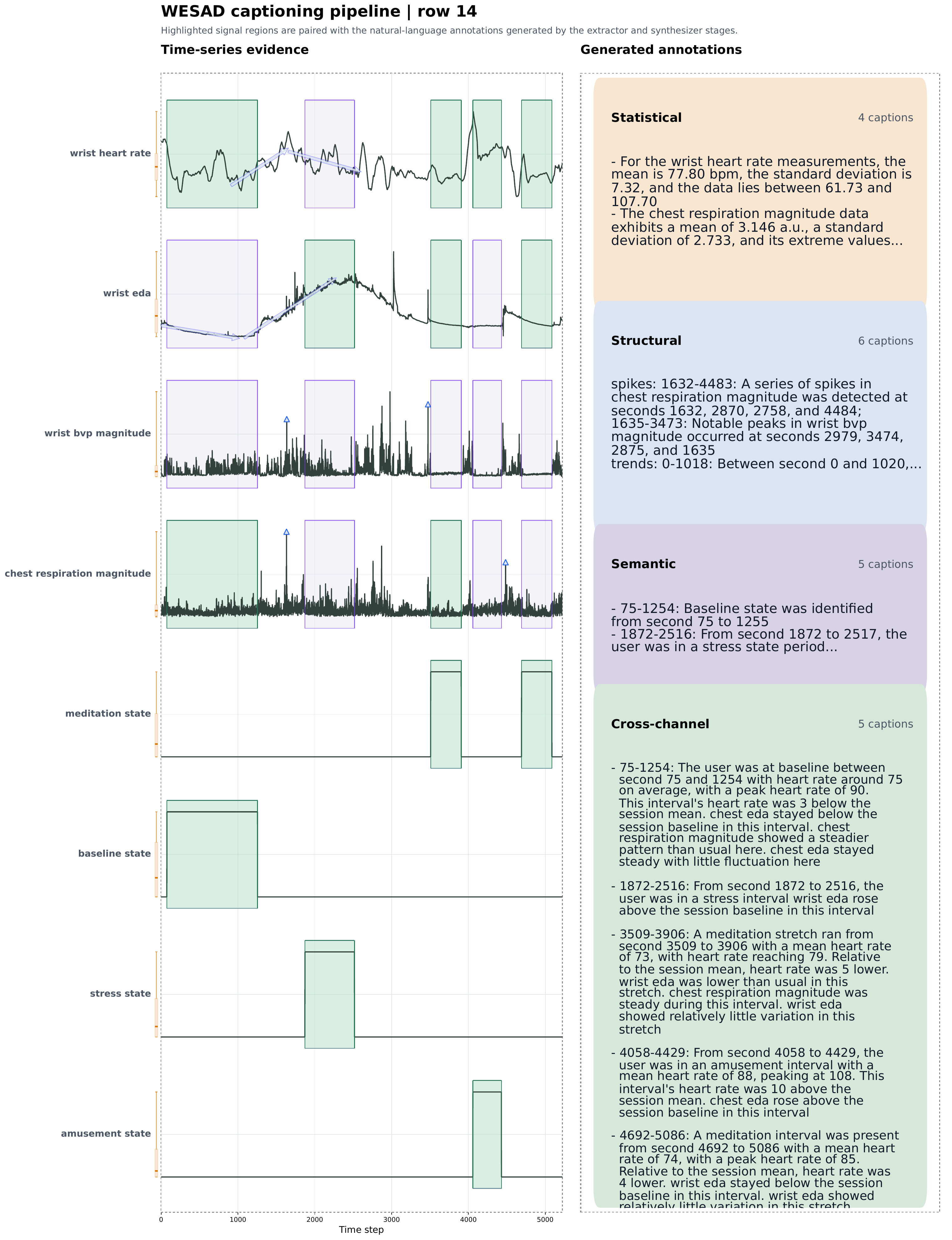}
    \caption{\textsc{WESADCap} subject 15.}
    \label{fig:WESADCap15}
\end{figure}

The captioned examples for \textsc{WESADCap} can be seen in Figures \ref{fig:WESADCap1}-\ref{fig:WESADCap15}.

\begin{figure}[!htbp]
    \centering
    \includegraphics[width=\linewidth,alt={WESAD subject 2 physiological and state traces with highlighted intervals paired with statistical, structural, semantic, and cross-channel captions.}]{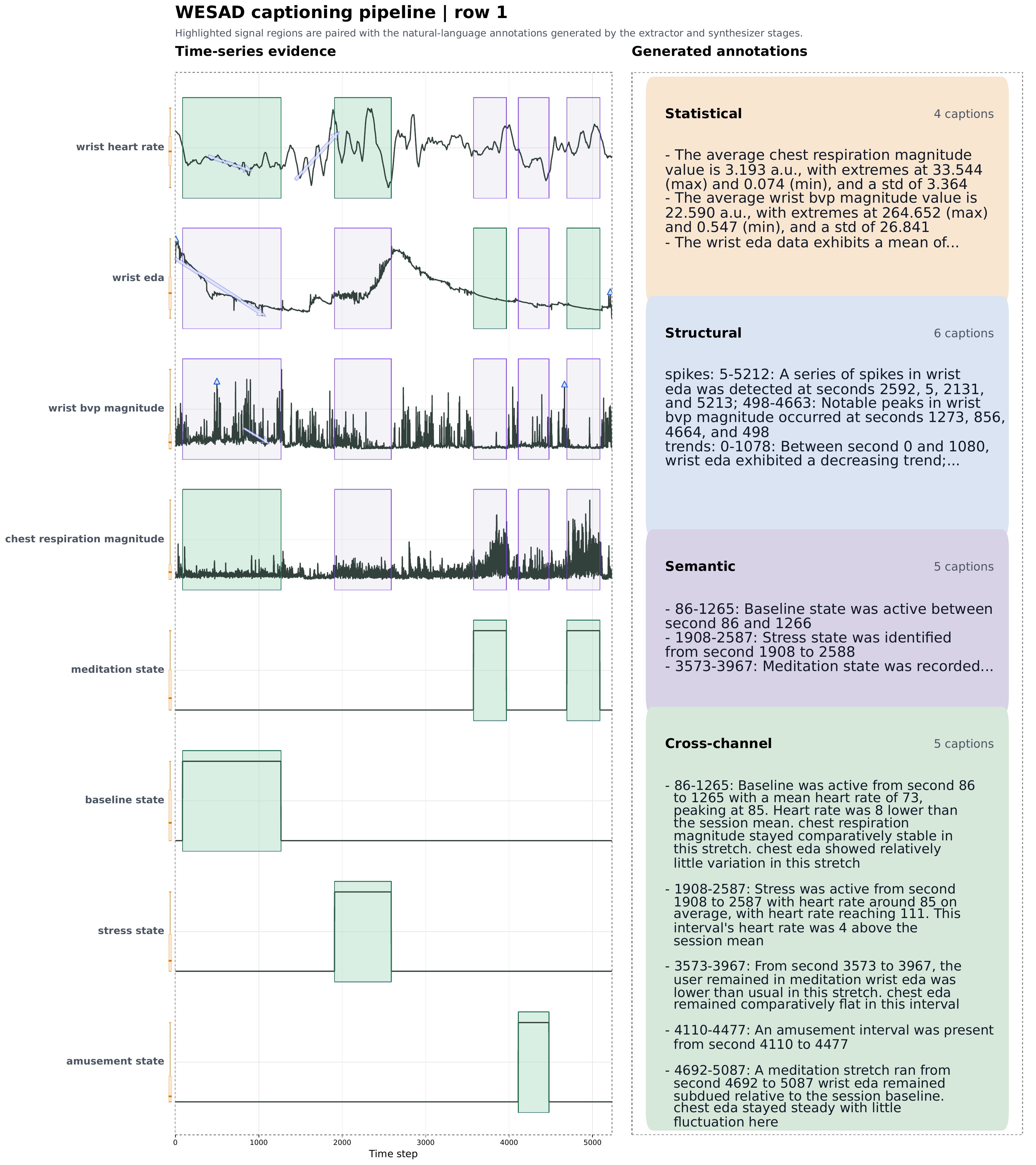}
    \caption{\textsc{WESADCap} subject 2.}
\end{figure}

\begin{figure}[!htbp]
    \centering
    \includegraphics[width=\linewidth,alt={WESAD subject 3 physiological and state traces with highlighted intervals paired with statistical, structural, semantic, and cross-channel captions.}]{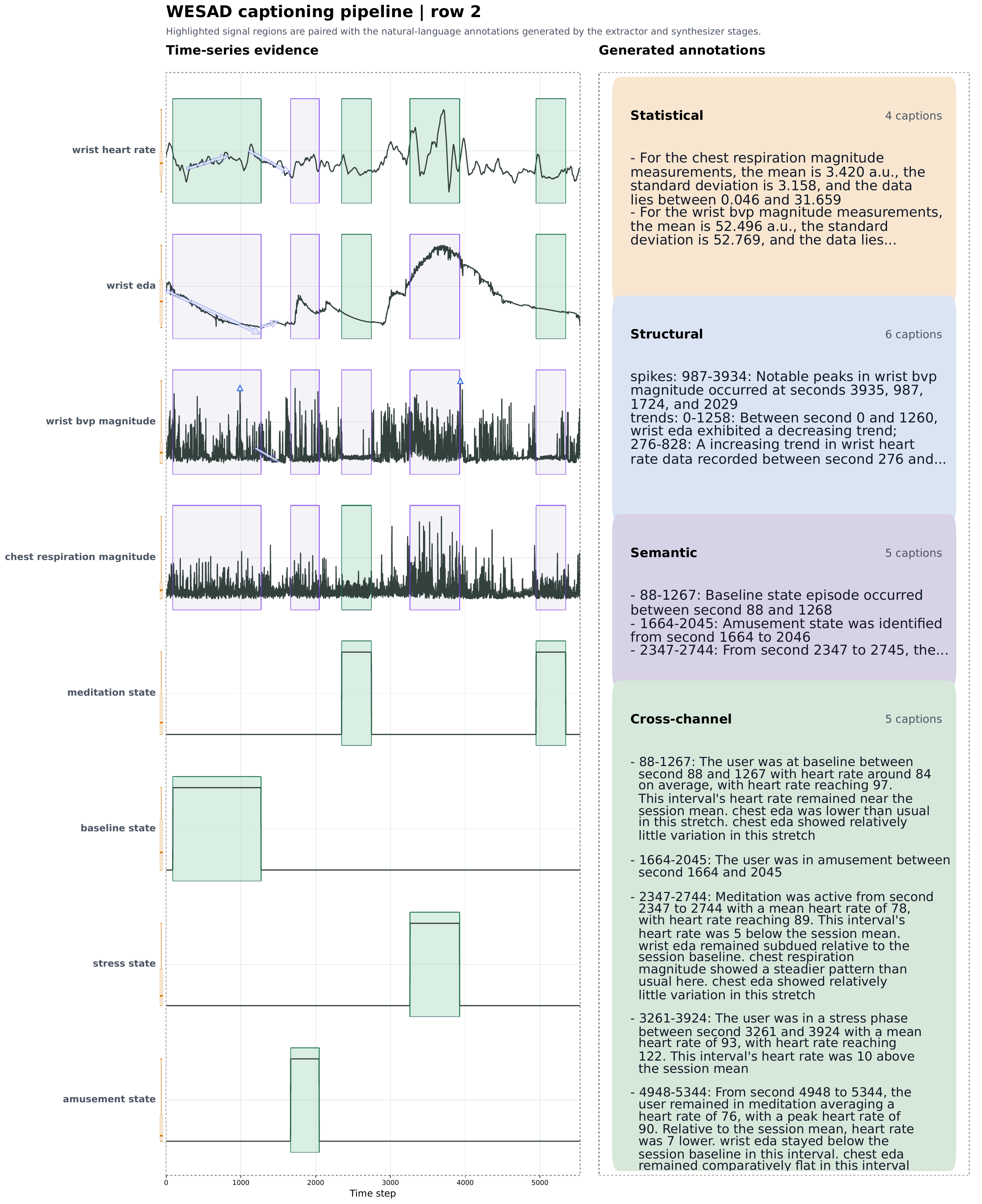}
    \caption{\textsc{WESADCap} subject 3.}
\end{figure}

\begin{figure}[!htbp]
    \centering
    \includegraphics[width=\linewidth,alt={WESAD subject 4 physiological and state traces with highlighted intervals paired with statistical, structural, semantic, and cross-channel captions.}]{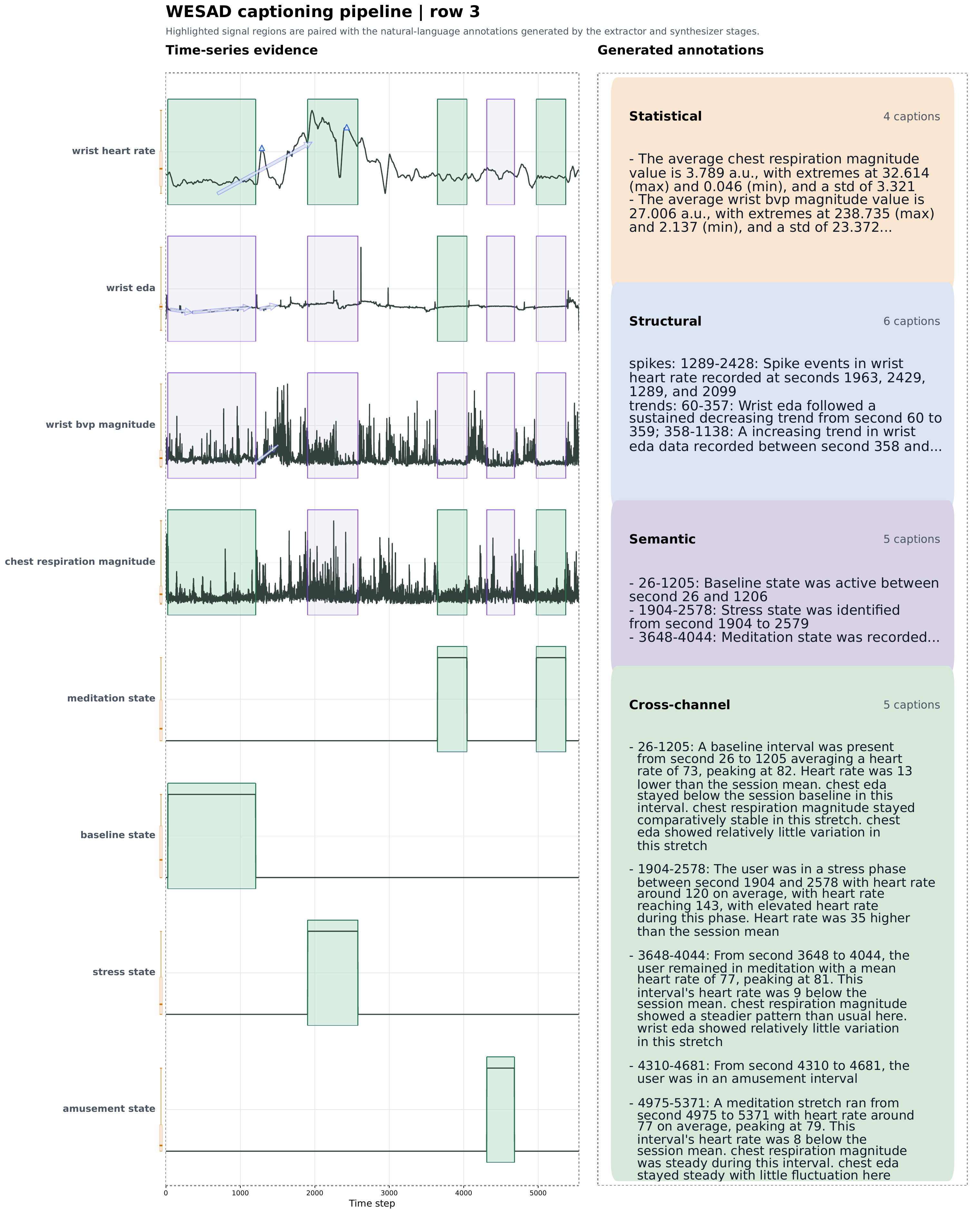}
    \caption{\textsc{WESADCap} subject 4.}
\end{figure}

\begin{figure}[!htbp]
    \centering
    \includegraphics[width=\linewidth,alt={WESAD subject 5 physiological and state traces with highlighted intervals paired with statistical, structural, semantic, and cross-channel captions.}]{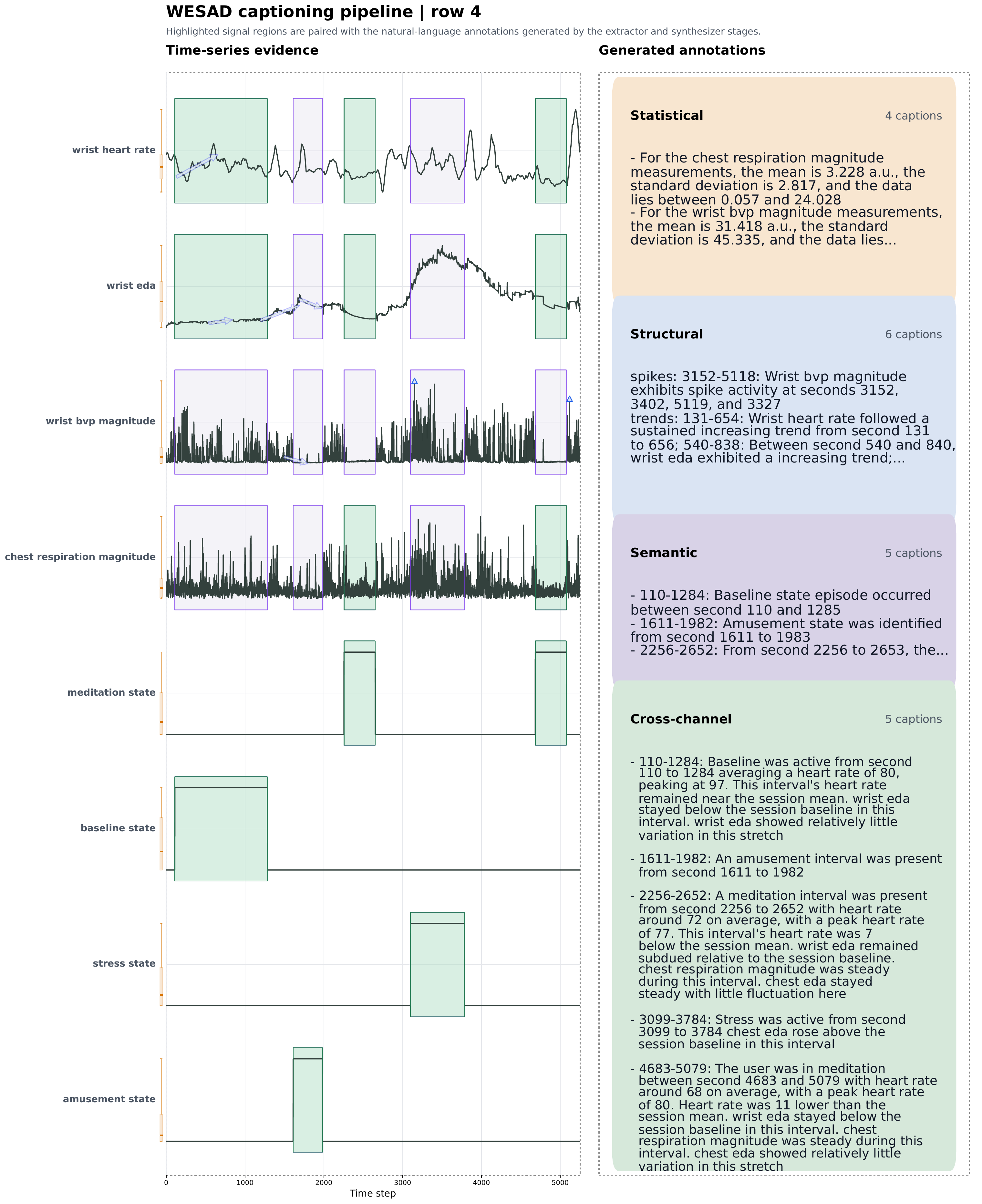}
    \caption{\textsc{WESADCap} subject 5.}
\end{figure}

\begin{figure}[!htbp]
    \centering
    \includegraphics[width=\linewidth,height=0.92\textheight,keepaspectratio,alt={WESAD subject 6 physiological and state traces with highlighted intervals paired with statistical, structural, semantic, and cross-channel captions.}]{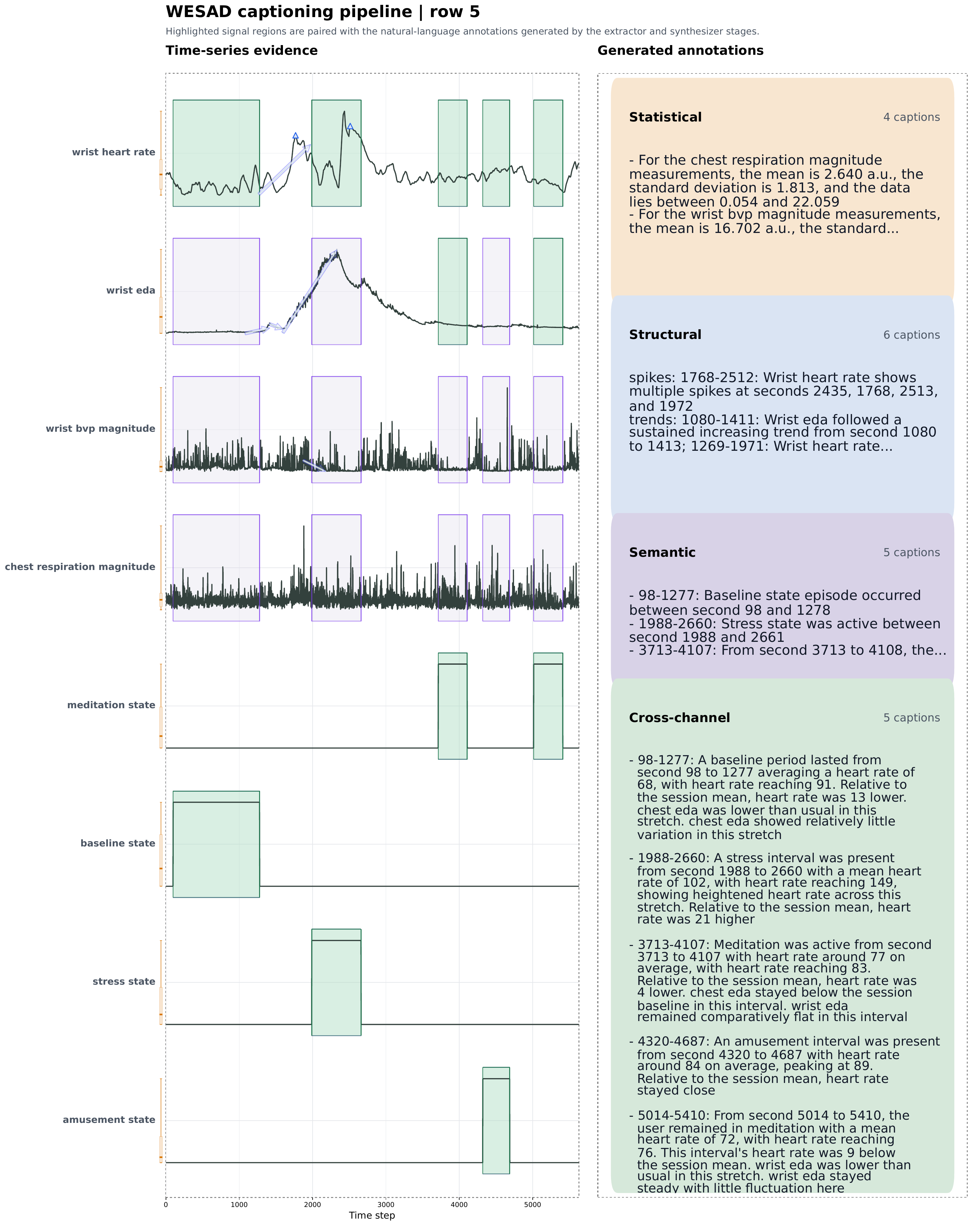}
    \caption{\textsc{WESADCap} subject 6.}
\end{figure}

\begin{figure}[!htbp]
    \centering
    \includegraphics[width=\linewidth,alt={WESAD subject 7 physiological and state traces with highlighted intervals paired with statistical, structural, semantic, and cross-channel captions.}]{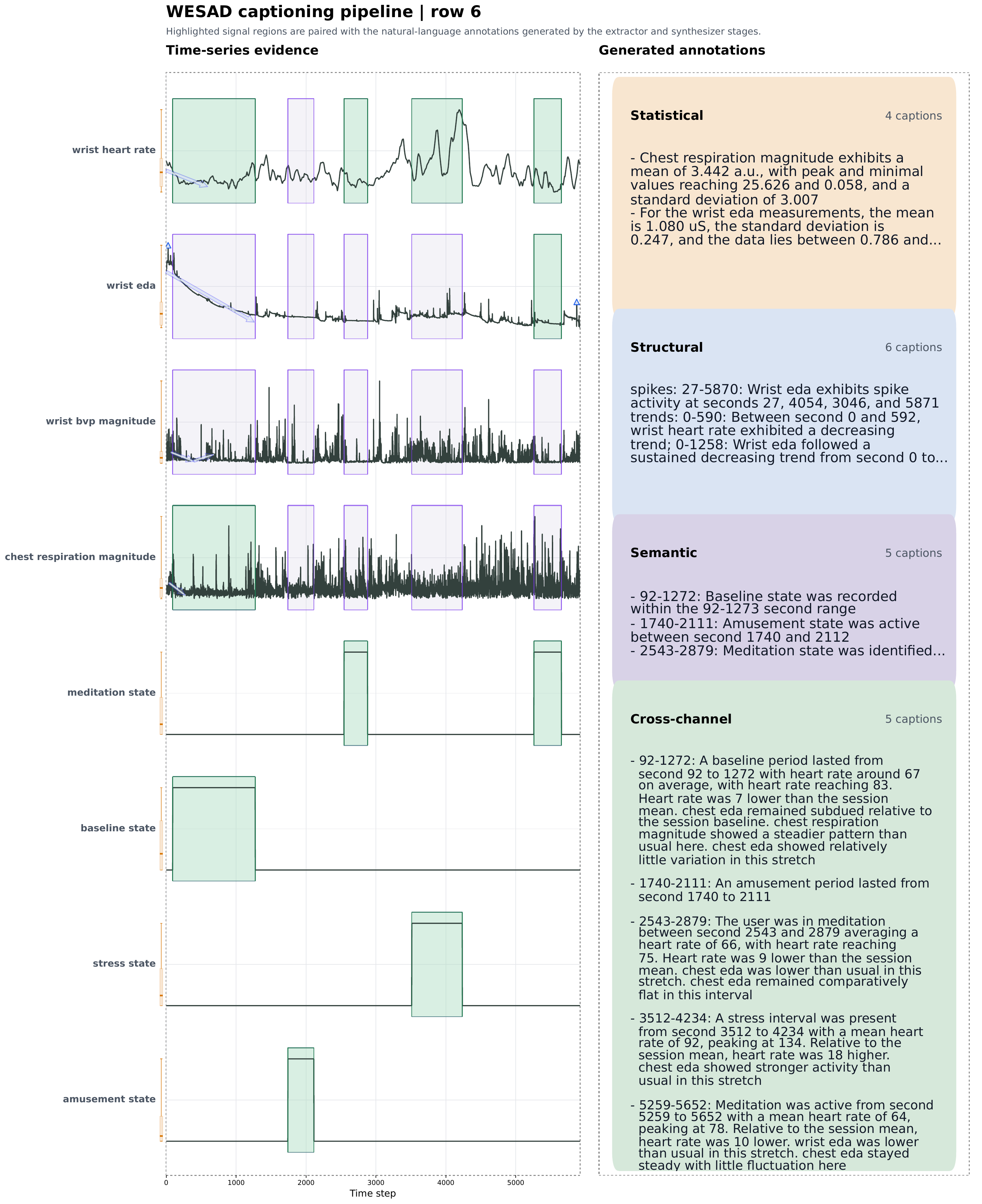}
    \caption{\textsc{WESADCap} subject 7.}
\end{figure}

\begin{figure}[!htbp]
    \centering
    \includegraphics[width=\linewidth,alt={WESAD subject 8 physiological and state traces with highlighted intervals paired with statistical, structural, semantic, and cross-channel captions.}]{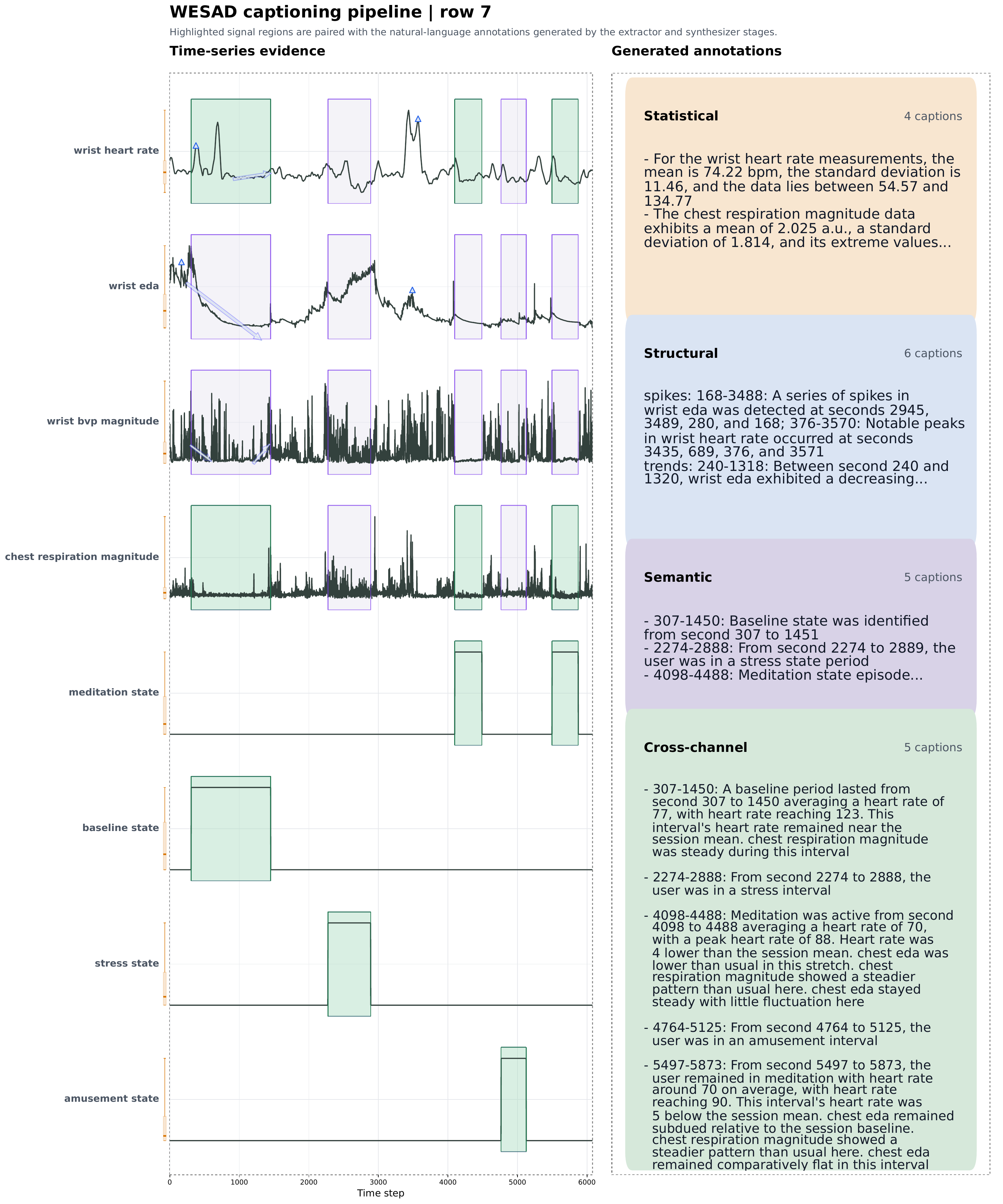}
    \caption{\textsc{WESADCap} subject 8.}
\end{figure}

\begin{figure}[!htbp]
    \centering
    \includegraphics[width=\linewidth,alt={WESAD subject 9 physiological and state traces with highlighted intervals paired with statistical, structural, semantic, and cross-channel captions.}]{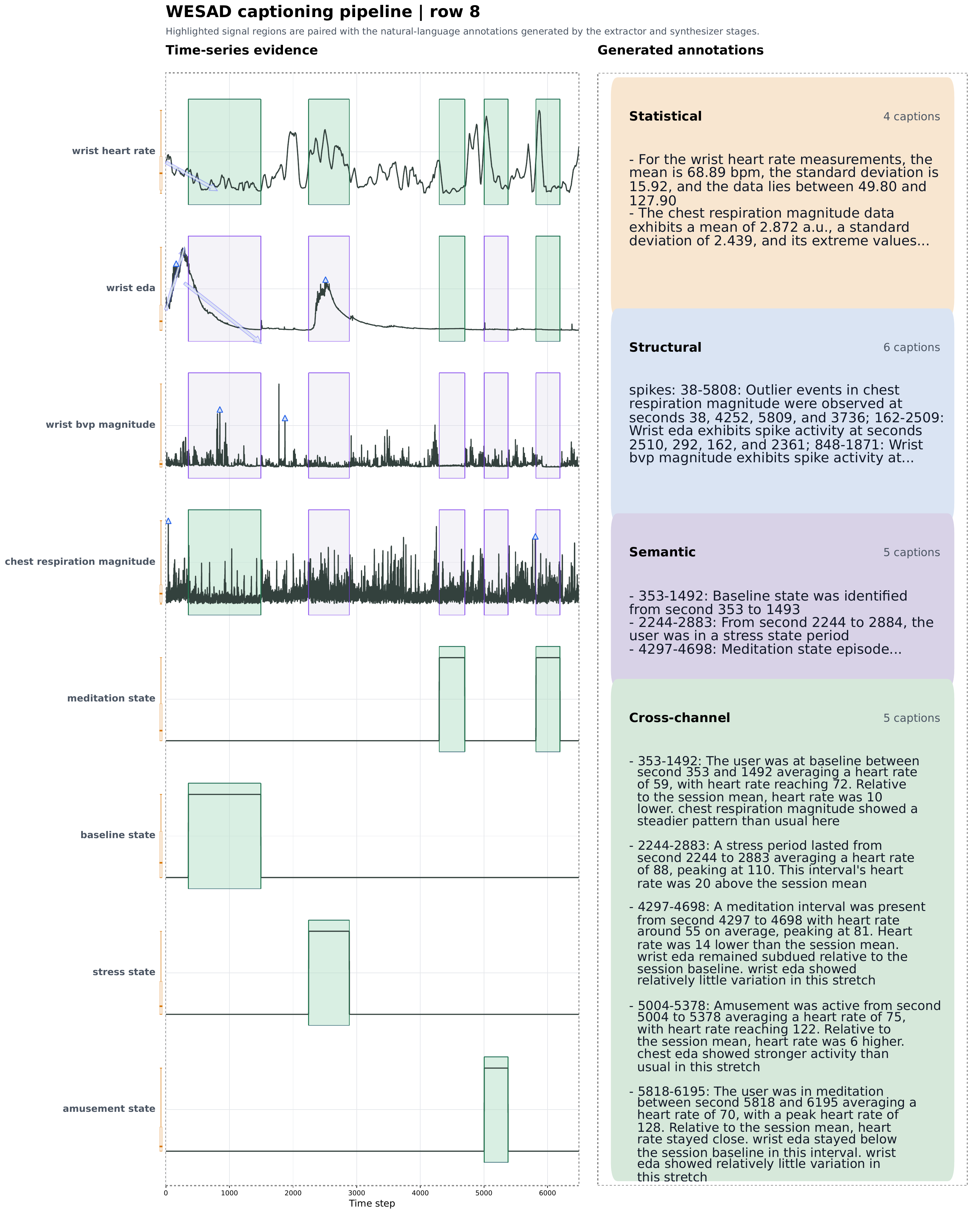}
    \caption{\textsc{WESADCap} subject 9.}
\end{figure}

\begin{figure}[!htbp]
    \centering
    \includegraphics[width=\linewidth,alt={WESAD subject 10 physiological and state traces with highlighted intervals paired with statistical, structural, semantic, and cross-channel captions.}]{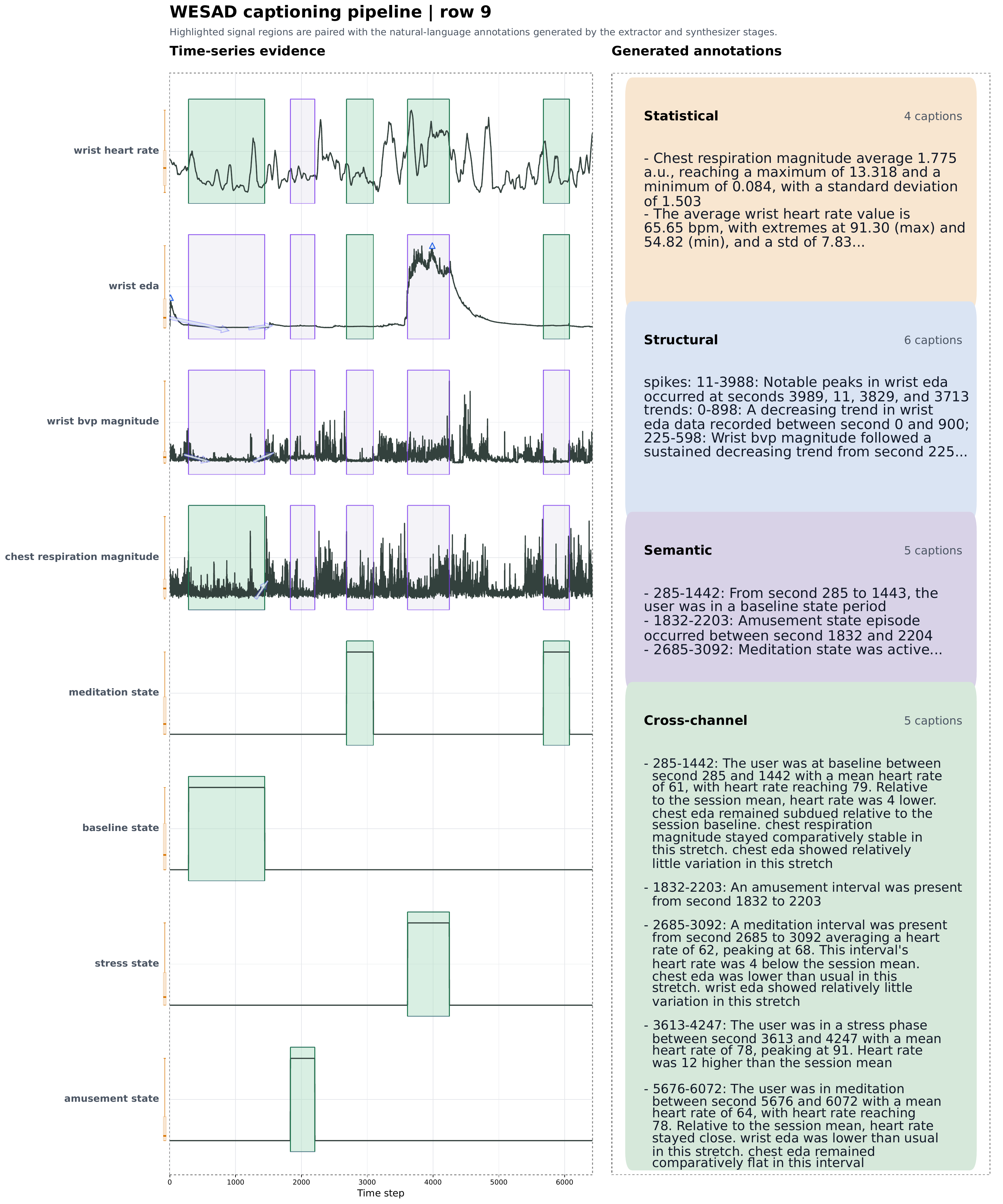}
    \caption{\textsc{WESADCap} subject 10.}
\end{figure}

\begin{figure}[!htbp]
    \centering
    \includegraphics[width=\linewidth,alt={WESAD subject 11 physiological and state traces with highlighted intervals paired with statistical, structural, semantic, and cross-channel captions.}]{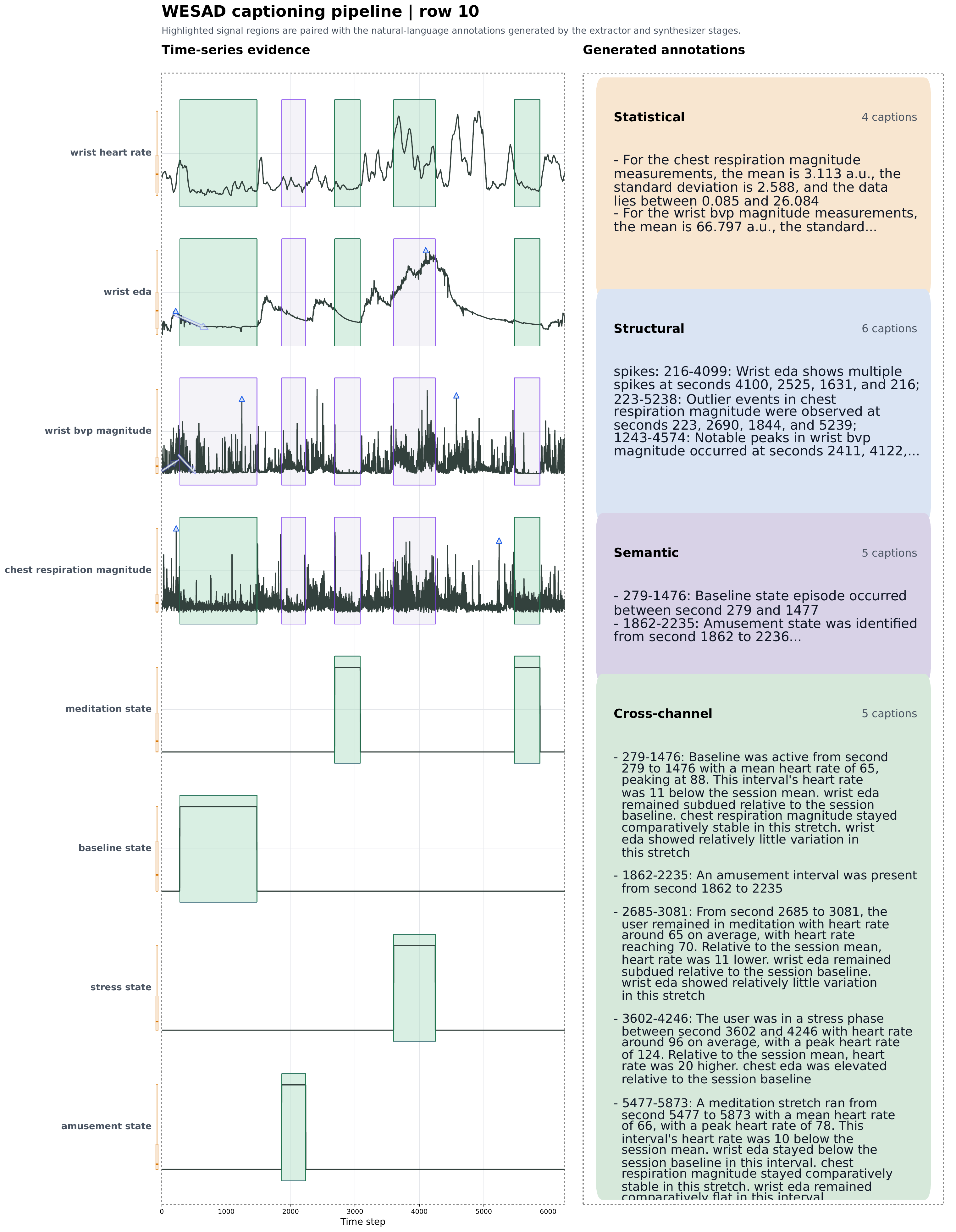}
    \caption{\textsc{WESADCap} subject 11.}
\end{figure}

\begin{figure}[!htbp]
    \centering
    \includegraphics[width=\linewidth,alt={WESAD subject 12 physiological and state traces with highlighted intervals paired with statistical, structural, semantic, and cross-channel captions.}]{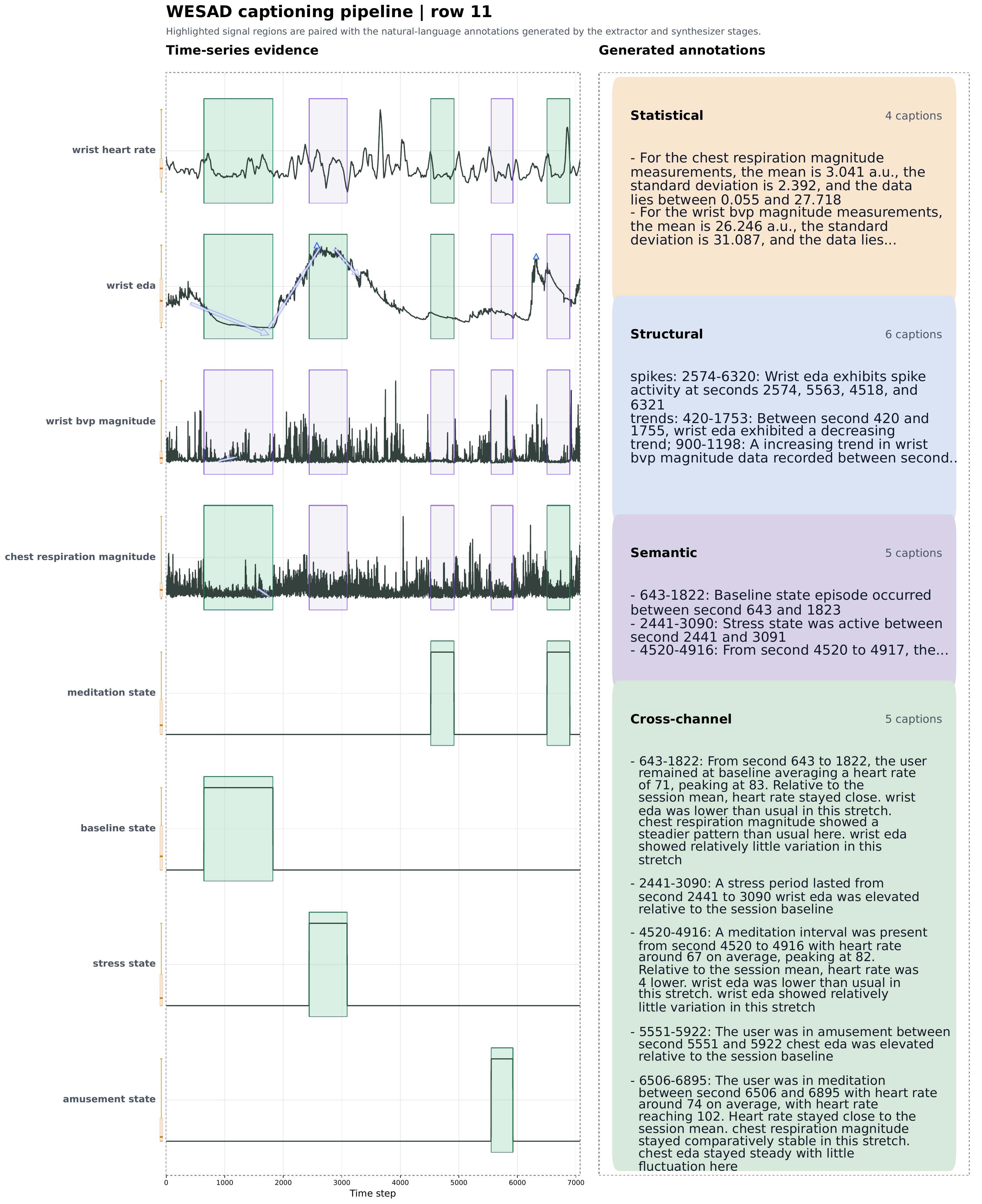}
    \caption{\textsc{WESADCap} subject 12.}
\end{figure}

\begin{figure}[!htbp]
    \centering
    \includegraphics[width=\linewidth,alt={WESAD subject 13 physiological and state traces with highlighted intervals paired with statistical, structural, semantic, and cross-channel captions.}]{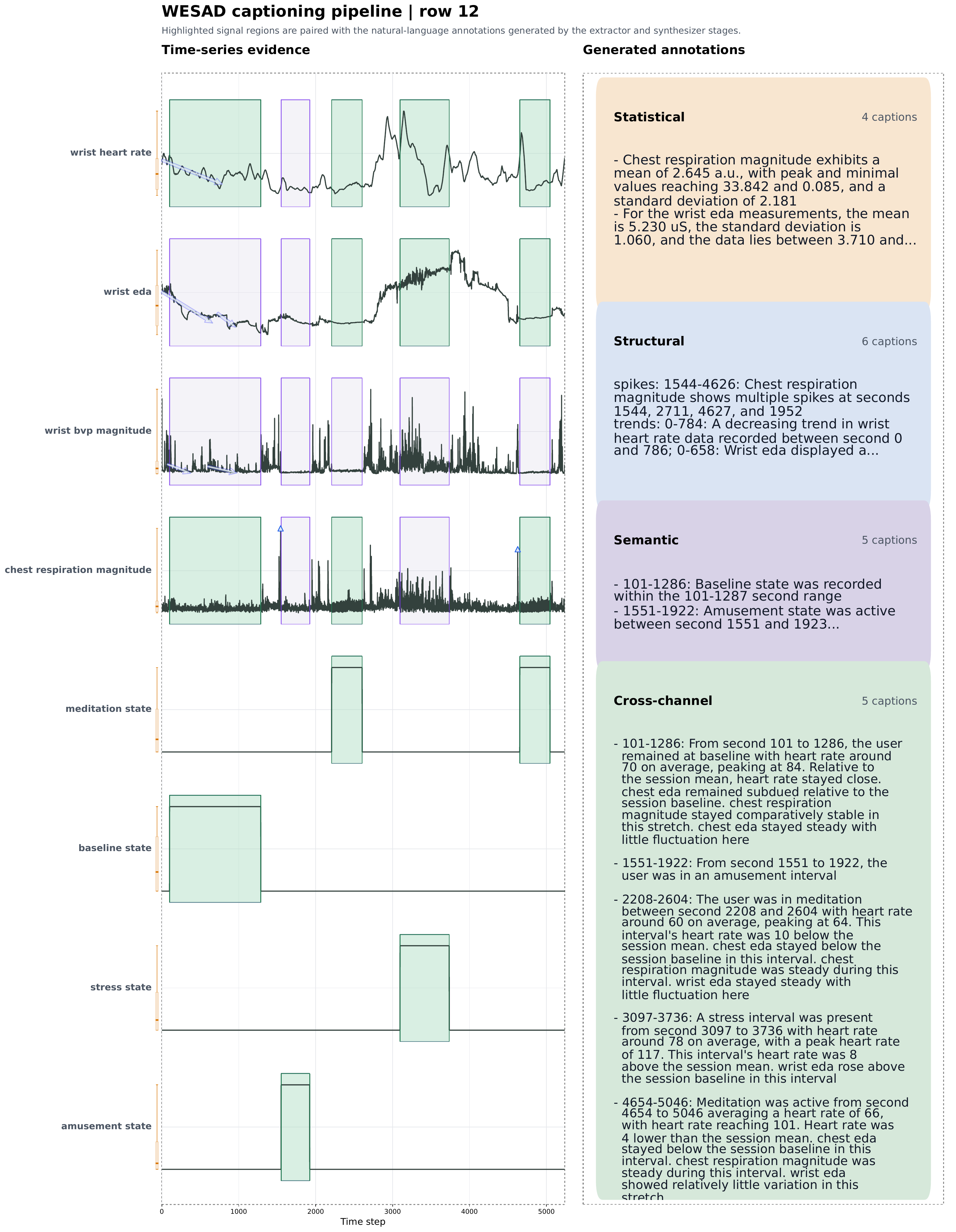}
    \caption{\textsc{WESADCap} subject 13.}
\end{figure}

\begin{figure}[!htbp]
    \centering
    \includegraphics[width=\linewidth,alt={WESAD subject 14 physiological and state traces with highlighted intervals paired with statistical, structural, semantic, and cross-channel captions.}]{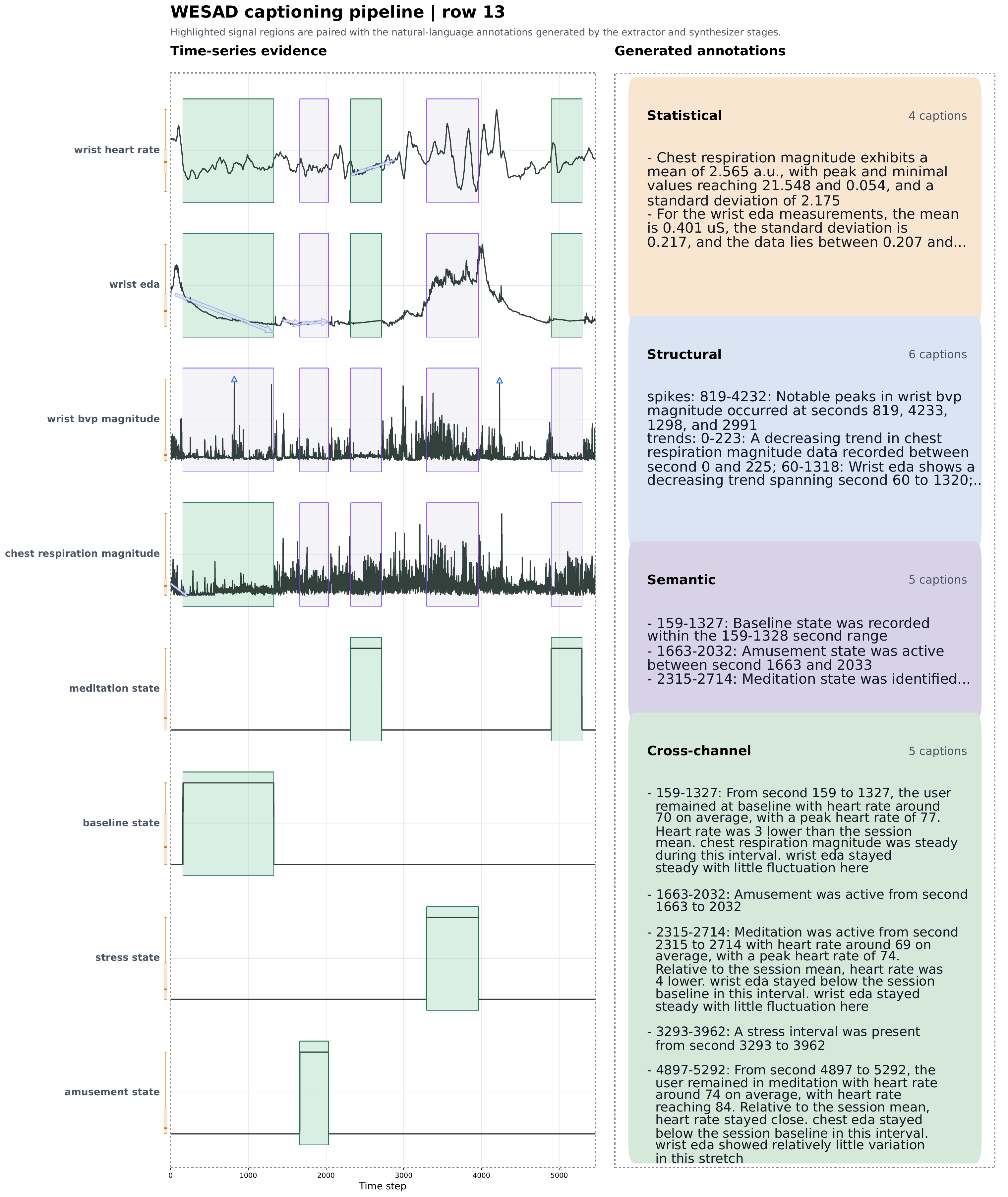}
    \caption{\textsc{WESADCap} subject 14.}
\end{figure}

\end{document}